\documentclass[runningheads]{llncs}
\usepackage{graphicx}

\usepackage{amsmath}

\usepackage{mathtools}
\usepackage[ruled, noline, linesnumbered]{algorithm2e} 
\usepackage{multirow}
\usepackage{tabularx}
\usepackage[dvipsnames,table,xcdraw,x11names]{xcolor}
\usepackage{float} %
\usepackage{booktabs} %
\usepackage{amsmath}
\usepackage{amssymb}
\usepackage{array}
\newcolumntype{P}[1]{>{\centering\arraybackslash}p{#1}}
\usepackage{adjustbox}
\newcommand*\samethanks[1][\value{footnote}]{\footnotemark[#1]}

\newcommand{\datasetName}{PFL-DocVQA\xspace}

\newcommand{\conf}{\mathit{conf}\xspace}
\newcommand{\memtest}{Memorization Test\xspace}
\newcommand{\sota}{SotA\xspace}

\newcommand{\vtfz}{VT5\textsubscript{0}\xspace}
\newcommand{\vtfc}{VT5\textsubscript{C}\xspace}
\newcommand{\vtffl}{VT5\textsubscript{FL}\xspace}
\newcommand{\vtfcdp}{VT5\textsubscript{C+DP}\xspace}
\newcommand{\vtffldp}{VT5\textsubscript{FL+DP}\xspace}

\newcommand{\Din}{\mathcal{D}_{\text{in}}\xspace}
\newcommand{\Dout}{\mathcal{D}_{\text{out}}\xspace}

\newcommand{\redin}{RED\textsubscript{$\mathcal{D}_{in}$}\xspace}
\newcommand{\redout}{RED\textsubscript{$\mathcal{D}_{out}$}\xspace}

\newcommand{\Ebudget}{{\large $\varepsilon$}}
\newcommand{\EbudgetS}{{\large $\varepsilon$}\xspace}

\newcommand{\gt}[1]{{\color{blue}#1}}

\usepackage{amsmath,amsfonts,bm}

\def\eqref#1{equation~\ref{#1}}

\def\1{\bm{1}}

\DeclareMathAlphabet{\mathsfit}{\encodingdefault}{\sfdefault}{m}{sl}
\SetMathAlphabet{\mathsfit}{bold}{\encodingdefault}{\sfdefault}{bx}{n}

\newcommand\mbf{\mathbf}
\usepackage[capitalize,noabbrev]{cleveref}

\newcommand{\Tgd}{T_{\mathsf{gd}}}

\begin{document}
\title{Privacy-Aware Document Visual Question Answering}
\author{Rubèn Tito\thanks{These authors contributed equally to this work.}\inst{1}\and
Khanh Nguyen\samethanks[1]\inst{1}\and
Marlon Tobaben\samethanks[1]\inst{2}\and
Raouf Kerkouche\inst{3}\and
Mohamed Ali Souibgui\inst{1}\and 
Kangsoo Jung\inst{4}\and
Joonas Jälkö\inst{2}\and
Vincent Poulain D'Andecy\inst{5}\and
Aurelie Joseph\inst{5}\and
Lei Kang\inst{1}\and
Ernest Valveny\inst{1}\and
Antti Honkela\inst{2}\and
Mario Fritz\inst{3}\and
Dimosthenis Karatzas\inst{1}
}
\authorrunning{R. Tito et al.}
\institute{Computer Vision Center, Universitat Autonoma de Barcelona, Spain \and University of Helsinki, Finland \and  CISPA Helmholtz Center for Information Security, Germany \and French Institute for Research in Computer Science and Automation (INRIA), France \and Yooz, France\\ 
\email{rperez@cvc.uab.cat} ~~~~~~~ \email{knguyen@cvc.uab.cat} ~~~~~~~ \email{marlon.tobaben@helsinki.fi}
}

\maketitle              %

\begin{abstract}
Document Visual Question Answering (DocVQA) has quickly grown into a central task of document understanding. 
But despite the fact that documents contain sensitive or copyrighted information, none of the current DocVQA methods offers strong privacy guarantees.

In this work, we explore privacy in the domain of DocVQA for the first time, highlighting privacy issues in state of the art multi-modal LLM models used for DocVQA, and explore possible solutions.

Specifically, we focus on invoice processing as a realistic document understanding  scenario, and propose a large scale DocVQA dataset comprising invoice documents and associated questions and answers. We employ a federated learning scheme, that reflects the real-life distribution of documents in different businesses, and we explore the use case where the data of the invoice provider is the sensitive information to be protected.

We demonstrate that non-private models tend to memorise, a behaviour that can lead to exposing private information. We then evaluate baseline training schemes employing federated learning and differential privacy in this multi-modal scenario, where the sensitive information might be exposed through either or both of the two input modalities: vision (document image) or language (OCR tokens).

Finally, we design attacks exploiting the memorisation effect of the model, and demonstrate their effectiveness in probing a representative DocVQA models.

\keywords{DocVQA  \and Federated Learning \and Differential Privacy}
\end{abstract}

\section{Introduction}

Automatic document processing enables the vast majority of daily interactions with and between institutions. From a research viewpoint, document understanding is a multi-modal endeavour combining reading systems and the visual analysis of document images with language processing and, more recently, language-based interaction. Document Visual Question Answering (DocVQA) was introduced in 2019 \cite{mathew2020document,mathew2021docvqa} and quickly reshaped the state of the art, by introducing specialised, large scale, multi-modal LLMs for document understanding \cite{xu2021layoutlmv2,huang2022layoutlmv3,powalski2021going}.

\begin{figure}
    \centering
    \begin{tabular}[t]{@{} p{0.35\linewidth} p{0.65\linewidth} @{}}
        \begin{minipage}[t]{\linewidth}
            \textbf{Question:} What is the \\ provider of this document? \\
            \textbf{Ground truth Answer:} \\\gt{WMTW} \\ \\
            \textbf{Non-private Model:} \\WMTW \\ 
            \textbf{Private Model:} \\Meredith Thompson 
        \end{minipage}
        &
        \begin{minipage}[t]{\linewidth}
            \vspace{-15pt} %
            \frame{\includegraphics[width=0.925\linewidth, height=4cm]{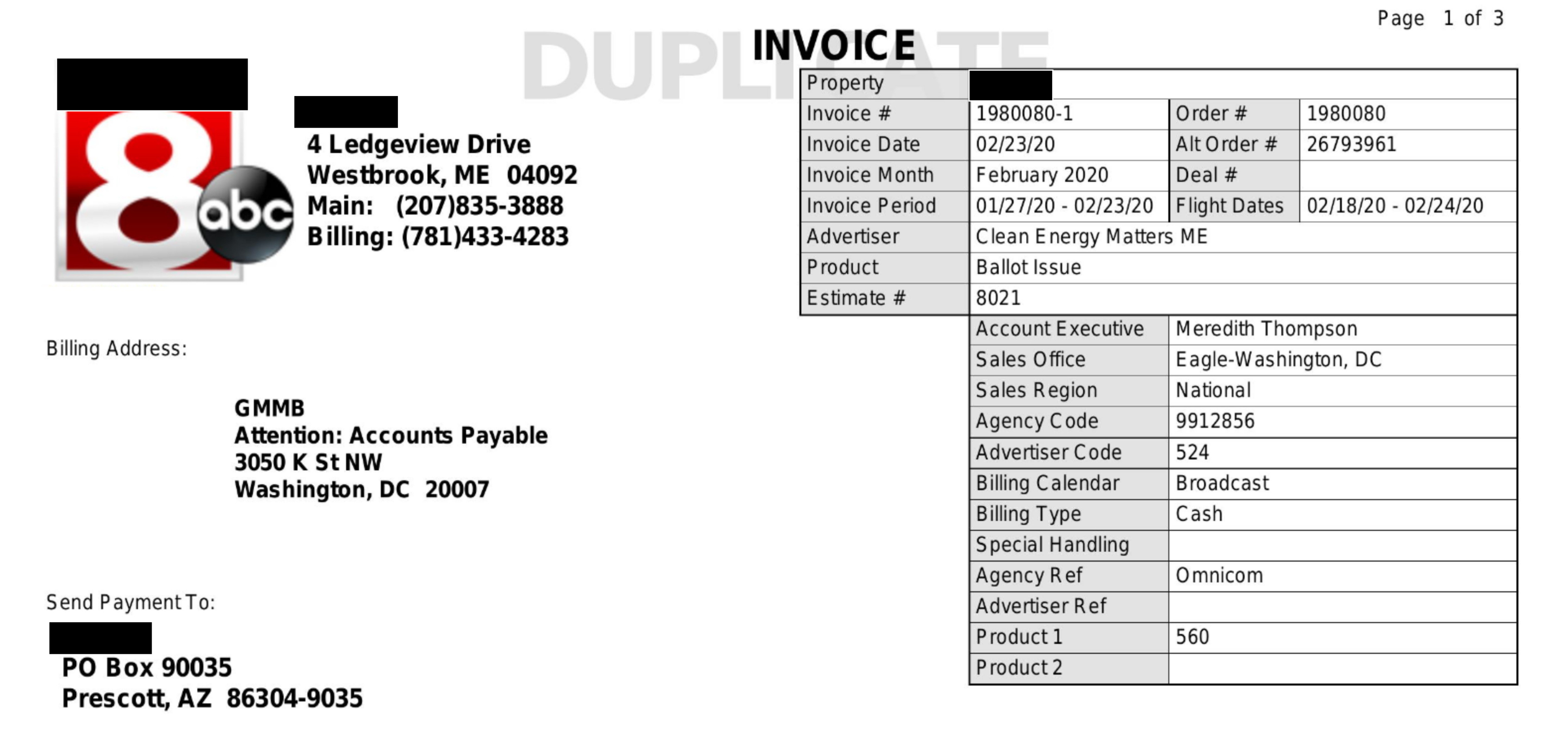}}
        \end{minipage} \\
    \end{tabular}
    \caption{The risk of malicious attacks on trained DocVQA models, such as exploiting memorization, is evident in the \datasetName dataset. 
     Adversaries can cue the model through the visual modality, to invoke memory and reveal sensitive information that is not explicitly in the document (e.g. in this example the provider's name).
     We show how this behaviour can be exploited to attack the models, and take first steps to mitigate the problem.
     }
     \label{fig:example_qas}
 \end{figure}

Despite the fact that documents contain sensitive or copyrighted information, none of the current DocVQA methods offers strong privacy guarantees. On the contrary, such models tend to memorise information, and often hallucinate responses drawing information from their training data, as illustrated in Fig \ref{fig:example_qas}.

Privacy-preserving methods have advanced considerably over the past decade with differential privacy (DP)~\cite{dwork2006calibrating} being the gold standard for protecting training data~\cite{kurakin2022toward,de2022unlocking,mehta2023large,cattan2022,tobaben2023Efficacy}, nevertheless the DocVQA scenario presents important challenges to overcome. On one hand, privacy-preserving methods are not specifically designed for multi-modal scenarios, where the sensitive information might be exposed through any of the input modalities, or a combination of them. On the other hand, the models employed for DocVQA tend to be large and cumbersome, and require long pre-training and fine-tuning stages.

Moreover, documents in real-life cannot be freely exchanged. More often than not, different entities have access to distinct sets of documents that cannot be shared due to legal reasons. As a result, most document analysis datasets tend to be small, or focus on non-realistic, out of copyright document collections. Collaborative learning approaches, that do not require centralising the training data, are a valid alternative that would allow exploiting real-life data; nevertheless, such methods are not currently used by the document analysis community.

In this work, we highlight privacy issues in state of the art multimodal LLM models used for DocVQA, and propose possible solutions. In addition, we employ a federated scheme for learning, that reflects the real-life distribution of documents.

We focus on invoice processing, as a real-life document understanding scenario.
Typically service providers issue invoices to clients, who use automated tools to extract information and take appropriate actions. In this scenario the sensitive information that needs to be protected is the invoice provider data contained in the training set. The trained model can potentially expose this information through vision (logo, layout of the invoice, colour scheme used, etc) and / or language (the provider's name, their vat number, address, telephone or any other identifying information in the textual part of the document). In this scenario, different clients receive invoices from distinct providers, highlighting the unbalanced and non i.i.d. distribution of data in the real world, in terms of invoice providers.

In this work we explore the application of privacy-preserving methods for fine-tuning a pre-trained state-of-the-art DocVQA model in a federated setting. For that, we create a new dataset where data is split into several groups (corresponding to clients in a real scenario), each one with a different distribution of providers. With this dataset we define two new tasks. The first one consists of collaborative learning of a single model among all the clients without sharing the training data. The second one aims at learning with privacy guarantees to protect the identity of the providers used during training.
Specifically, our contributions are:
\begin{itemize}
    \item We put forward a new dataset for private federated DocVQA, focused on the real-life scenario of invoice processing.
    \item We demonstrate that state-of-the-art non-private models exhibit a memorisation effect, that can lead to exposing sensitive information, and can be used to attack the model.
    \item We design a series of attacks exploiting the memorisation behaviour of the models.
    \item We evaluate different training methods employing Federated Learning (FL) and  Differential Privacy (DP) to train a \sota DocVQA model with privacy guarantees. We evaluate these methods using our proposed attacks, and show that FL+DP can mitigate privacy issues to certain degree.
\end{itemize}

\section{Related Work}

\noindent \textbf{Document Visual Question Answering.}
DocVQA has gained a lot of attention as a unified approach based on answering natural language questions to extract information from documents. Consequently, many datasets are nowadays available tackling different domains and challenges such as industry documents~\cite{mathew2020document,mathew2021docvqa,tito2021icdar,tito2023hierarchical}, infographics~\cite{mathew2022infographicvqa}, multidomain~\cite{van2023document,van2023icdar}, open-ended questions~\cite{tanaka2021visualmrc}, multilingual~\cite{qi2022dureadervis}, multipage~\cite{tito2023hierarchical} or collections of documents~\cite{tito2021document}. Despite the existence of many small and medium sized datasets one of the major challenges is still the lack of a large scale generic dataset that can be used for multiple scenarios. 
One of the main reasons for that is the %
sensitive content of many documents and the copyright issues they are subject to, that prevent most document holders from publicly releasing their data. In this direction, federated learning techniques coupled with privacy-preserving methods can offer a way to facilitate the use of distributed and/or private datasets among different entities.

Existing models for DocVQA have evolved from text-only span-prediction models \cite{devlin2018bert}, to multimodal methods that leverage different self-supervised pretraining tasks to align all the modalities over large scale datasets, either using OCR and layout information \cite{huang2022layoutlmv3,powalski2021going} or working directly with the input image without applying OCR~\cite{Lee23Pix2Struct,Kim22Donut}. Recently, some methods~\cite{Wang23LatinPrompt,Ye23mPlugDoc} have also tried to leverage  the zero-shot capabilities of pre-trained LLMs by adding an encoder to extract visual features from the document image and use these features and question as input to the LLM, fine-tuned via instruction-tuning. %
Since non-OCR methods show lower performance than OCR-based ones, and LLM-based methods are costly to fine-tune, in this work we will use VisualT5 (VT5), a multimodal generative method that has a strong performance on the  DocVQA task, while it is easy to fine-tune allowing us to focus on federated learning and privacy-preserving techniques on multimodal data.

\noindent \textbf{Differentially Private Federated Learning.}
Federated Learning (FL)~\cite{McMahan2017fedavg,shokri2015privacy}, allows collaborative model training among several entities (also known as clients) without sharing the entire data set of a client. Instead, only the trained model and its updates are shared between a central server and the different clients. The central server aggregates the model updates from all the clients while the data is kept locally at the client side. 
Even though federated learning is more private than the centralized approach, many attacks have shown that a significant amount of information can still be inferred from the updates/models shared between the clients and the server during training or afterwards~\cite{Property,nasr2019comprehensive,DLG,IDLG,VFLllgusenix,LLG,li2021label,NEURIPS2020_c4ede56b,li2022auditing}.
Adversaries can be either a participant, the server, or an external entity with access to the released trained model.

Among these attacks, Membership Inference Attacks (MIA)s~\cite{nasr2019comprehensive,shokri2017membership,carlini2022membership} — which aim to infer whether a specific record is included in a dataset — are among the most widely used methods for assessing privacy risks.
Recently, MIAs for multi-modal models \cite{hu2022m,ko2023practical} have been proposed. Both attacks use a pre-trained image-text model to identify matching pairs, but these attacks are not applicable to our DocVQA setting. In our work we aim to identify group membership (whether any invoice from a specific provider is included in the training set) in contrast to identifying specific records (specific invoices in our case). Unlike previous studies on MIAs, which rely on training datasets to evaluate attack models, our attack model uses only data sourced from the target group's distribution to test the attack model. This approach enables a more realistic evaluation of MIAs.

In order to mitigate privacy attacks, Differential Privacy (DP)~\cite{dwork2006calibrating} has emerged as the gold standard for formalizing privacy guarantees~\cite{kurakin2022toward,de2022unlocking,mehta2023large,cattan2022,tobaben2023Efficacy}. DP provides protection against membership inference attacks as it ensures that there is a high probability that a machine learning model trained under DP is similar whether or not the data of a particular entity (provider in our case) has been used for training the model.

However, DP introduces a trade-off between utility and privacy as model training under DP requires clipping updates and adding random noise, which have an impact on the accuracy of the model. The current \sota approaches for training large models with high utility under DP rely on transfer learning and utilize models pretrained on large public datasets that are then fine-tuned \cite{yosinski2014transferable} on private datasets. These approaches have been shown to be effective in multiple domains where large public datasets are available such as NLP~\cite{li2022large,yu2022differentially} and computer vision~\cite{kurakin2022toward,de2022unlocking,mehta2023large,cattan2022} even when the private dataset is small~\cite{tobaben2023Efficacy,tobaben2024understanding}. 
Parameter-efficient fine-tuning~\cite{houlsby2019parameter} using adaptors such as LoRA~\cite{hu2022LoRA} has been shown to be competitive in (federated) transfer learning under DP~\cite{yu2022differentially,tobaben2023Efficacy}. Alternatively, compression has been proposed to train better DP models \cite{FL_CS_DP,FL_Cons_DP}.

In this work, we delve into the application of privacy-preserving techniques for fine-tuning a state-of-the-art Document Visual Question Answering (DocVQA) model within a federated learning framework. Our key contributions include the introduction of the first dataset tailored for private federated learning in DocVQA, evidence of a memorization effect in state-of-the-art non-private models that could compromise sensitive information, the development of attack strategies exploiting this memorization, and the assessment of various training methodologies incorporating Federated Learning and DP. Through this evaluation, employing our proposed attacks, we demonstrate that combining Federated Learning with DP can effectively mitigate privacy risks to a significant extent.

\section{\datasetName Dataset} \label{sec:dataset}

In real-life scenarios, the free exchange of documents is frequently restricted. Indeed, various entities possess unique sets of documents that cannot be shared due to privacy constraints. Consequently, datasets for document analysis often tend to be limited in size or concentrate on unrealistic, out-of-copyright collections of documents. In this context, collaborative learning approaches, which do not necessitate the centralization of training data, emerge as a viable alternative for leveraging real-life data. However, these methods have not yet been adopted within the document analysis community.

The \datasetName dataset is designed to perform DocVQA in a federated learning and differential privacy set-up.  It is the first dataset for privacy-aware DocVQA that aims to expose privacy leakage issues in a realistic scenario. 
\datasetName  comprises invoice document images along with their OCR transcriptions and a set of question/answer pairs. In our setting, the sensitive data that needs to be protected is the invoice provider identity, regardless of the information used to discover it (provider name, email, address, logo...). %
Therefore, a malicious attack should not be able to exploit the trained model to reveal sensitive provider information or to discover if invoices from a particular provider were included in the training set.

\begin{figure}[h!]
    \centering
    \includegraphics[width=0.7\linewidth]{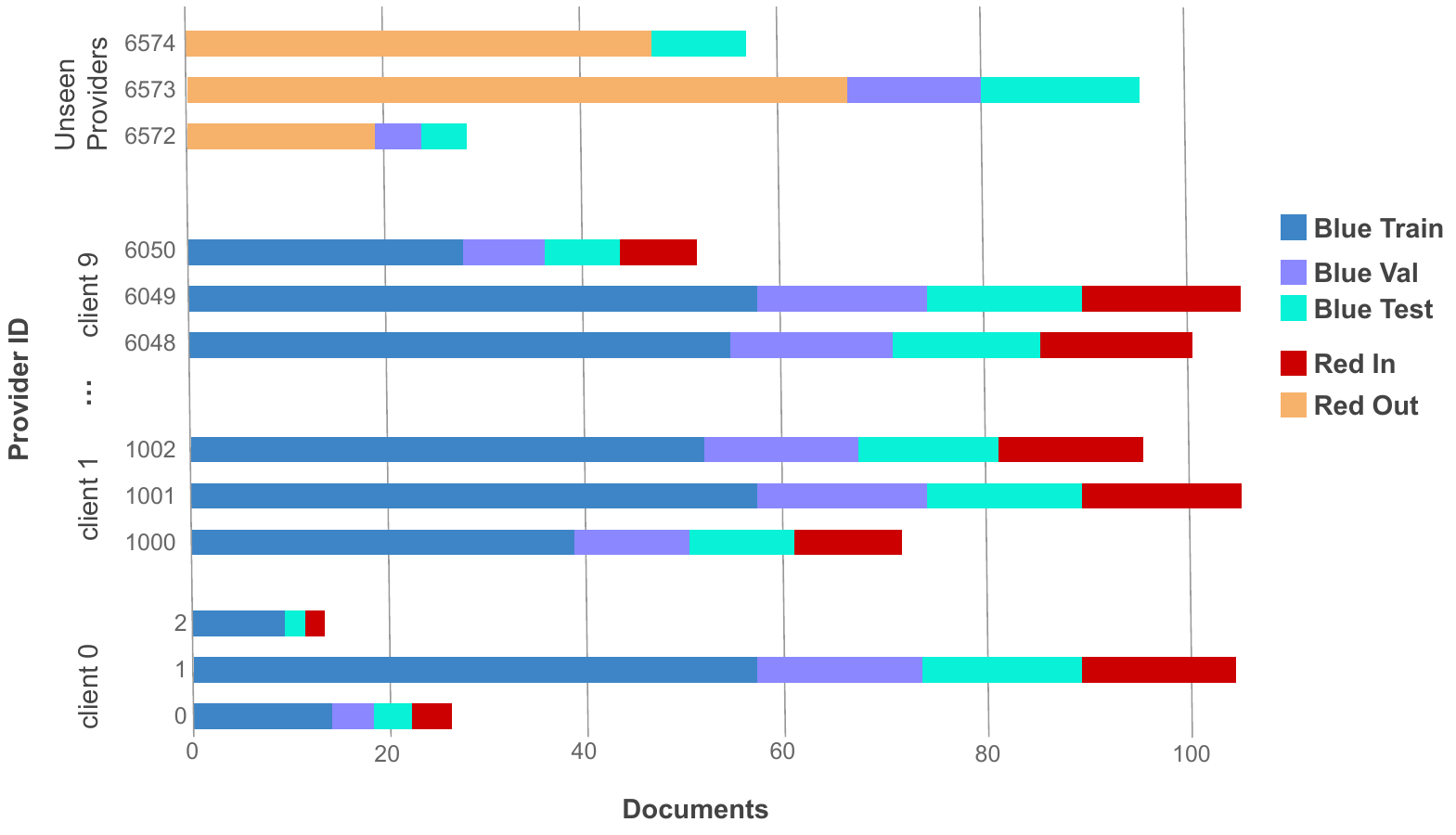}
    \caption{Distribution of providers and documents across different groups and splits. Every bar represents a specific provider, which contains a set of documents. The BLUE dataset is used for training the models, while the RED data is used for the attacks.}
    \label{fig:data_splits}
\end{figure}

The current version of the \datasetName dataset contains a total of $336,842$ question-answer pairs framed on $117,661$ pages of $37,669$ documents from $6,574$ different invoice providers.
The document images used in the dataset are sourced from the DocILE dataset~\cite{vsimsa2023docile}, which is designed for Key Information Localization and Extraction (KILE). The DocILE dataset comprises $939,147$ real business documents from public data sources \cite{ucsf,pif}, out of which $6,680$ are annotated with key-value pairs (i.e., a unique identifier (key) associated with a corresponding piece of data (value) within the document). We verified the annotations of the DocILE dataset to avoid OCR errors in the provider name 
and omitted samples that did not specify the provider's name. This resulted in using $4,968$ of the DocILE annotated documents. We extended this annotated set by labeling an additional set of $32,701$ documents. The key-value annotation of the invoices was done through Amazon Key-Information extraction tool \cite{amazon2023textract}, and was manually verified to guarantee its accuracy. Finally, we grouped documents by provider in a semi-automatic way.

The key-value pairs were used to construct the questions and answers of the \datasetName dataset. Questions are formed to inquire about the key, and the answers are the corresponding values. Questions are generated semi-automatically, by defining multiple templates for each key, and further rephrasing them using a LLM~\cite{openai2023gpt4} to achieve linguistic variability. %
For each key-value pair in the dataset, we randomly select one template to create the final question-answer pairs.

\datasetName is used in two tasks, thus, it is composed of two parts, or sub-datasets.
Primarily, in the first task, for training and evaluating machine learning privacy-preserving solutions on  DocVQA in a federated learning fashion. For this,  a base sub-set of \datasetName is used. Additionally, in the second task, membership inference attacks are designed to assess the privacy guarantees of the DocVQA models that were trained with the base data. These attacking approaches utilize an extension subset from the  \datasetName dataset. 
We refer to the base dataset as the ``BLUE'' 
dataset and to the extension dataset as the ``RED'' 
dataset. 
To construct the RED and BLUE datasets we split the providers into an \emph{in} set $\Din$ and an \emph{out} set $\Dout$, where $\Din$ are the providers seen during training and $\Dout$ are the providers that were not seen.

The BLUE data %
consists of a training set that is divided among $N$ clients (in our case we use $N=10$), a validation set and a test set. 
The training set of each of the $N$ clients contains invoices sampled from a different subset of $\Din$ providers, resulting in a highly non i.i.d. distribution. In the
 BLUE %
validation and test sets, we include documents from both $\Din$ and $\Dout$ providers.

The RED dataset is created by selecting half of its documents as $\Din$  providers (\redin), i.e.\ documents from providers that appear in the BLUE training data. The other half of the RED data consists of documents from  $\Dout$ providers (\redout), i.e.\ documents from providers not used for the BLUE training data. RED data is split into a training and test set. 
The different sets and clients of \datasetName are illustrated in Figure~\ref{fig:data_splits}. We also provide more details in the supplementary material.

\section{Visual T5 Model} \label{sec:vt5_model}

As a baseline for the DocVQA task we make use of a multimodal generative model named Visual T5 (VT5), a simplified version of the Hi-VT5~\cite{tito2023hierarchical} that leverages well-known existing state-of-the-art methods. While the adoption of a multimodal approach allows us to exploit the different modalities, it introduces the challenge of safeguarding private information across these modalities. Additionally, as a generative method, VT5 can produce a wide range of text, not limiting responses to predefined categories, thus making the preservation of sensitive information more challenging.

More specifically, the backbone architecture is based on T5~\cite{raffel2020exploring} augmented with spatial information, and visual features from document images encoded with DiT~\cite{li2022dit}. As shown in the supplement in \cref{sec:appendix_vt5}, the tokenized question, OCR tokens and encoded image patches are concatenated and fed into the encoder-decoder model to generate the answer following an autoregressive mechanism.

The T5 backbone is initialized with %
pretrained weights on the C4 dataset~\cite{raffel2020exploring} and the visual DiT is initialized with pre-trained weights for document classification. %
Then, we fine-tune the model on the Single Page DocVQA (SP-DocVQA) dataset~\cite{mathew2020document,mathew2021docvqa} for ten epochs. We call this model %
\emph{zero-shot baseline}: \vtfz.

To evaluate the DocVQA performance, we use two metrics: the standard accuracy (ACC) and ANLS~\cite{biten2019icdar,biten2019scene}. ANLS is a soft accuracy metric introduced to better deal with noisy answers due to OCR errors. %

\section{Provider Membership Inference Attack}
In this section we emphasize the privacy risks of the proposed DocVQA task in the real-world scenario where  models have access to sensitive information. First, we demonstrate the vulnerability of the centralized non-private model \vtfc to leak private training data through memorizing specific information of the provider, where overfitting is known to play a relevant role. Then, we propose different approaches to attack the model privacy via provider membership inference, that aims at differentiating the models' behavior between providers seen and not seen during training.

\subsection{Memorization Test}
\label{sec:memorization}

Table \ref{table:overfitting} shows the performance of the zero-shot baseline \vtfz and the centralised non-DP model \vtfc on the RED test set. \vtfz shows no performance difference between the \redin (documents from providers in $\Din$), and the \redout(documents from providers in $\Dout$). At the same time, \vtfc shows considerable performance difference, indicating that the model overfits to a certain degree when trained on the \datasetName dataset. We hypothesize that this is caused mainly by memorization.

\begin{table}[ht]
\caption{\textbf{DocVQA Performance of Non-Private Models on RED.} Metrics are reported in percentage. $\Delta$ indicates the difference between \redin and \redout.}\label{table:overfitting}
\begin{center}
\begin{tabular}{|c|cc|cc|cc|c|c|}
\hline
\multirow{2}{*}{Model} & \multicolumn{2}{c|}{RED}           & \multicolumn{2}{c|}{\redin}  & \multicolumn{2}{c|}{\redout}  & \multirow{2}{*}{$\Delta$ACC} & \multirow{2}{*}{$\Delta$ANLS} \\ \cline{2-7}
                       & \multicolumn{1}{c|}{ACC}   & ANLS  & \multicolumn{1}{c|}{ACC}   & ANLS  & \multicolumn{1}{c|}{ACC}   & ANLS  &                                &                                 \\ \hline
\vtfz   & \multicolumn{1}{c|}{37.72} & 43.66 & \multicolumn{1}{c|}{37.62} & 44.10  & \multicolumn{1}{c|}{37.84} & 43.18 & 0.22                           & 0.92                            \\
\vtfc   & \multicolumn{1}{c|}{81.40}  & 90.17 & \multicolumn{1}{c|}{85.92} & 93.68 & \multicolumn{1}{c|}{76.53} & 86.48 & \textbf{\textcolor{red}{9.39}}                           & \textbf{\textcolor{red}{7.20}}                             \\ \hline
\end{tabular}
\end{center}
\end{table}
\setlength{\tabcolsep}{1.4pt}

There have been extensive works \cite{carlini2022quantifying,ippolito2022preventing,tirumala2022memorization} that study the memorization of training data for language models, yet this behavior of such models in multi-modal settings remains under-explored. We thus perform a series of experiments on the RED data to demonstrate the existence of memorized information after fine-tuning on PFL-DocVQA.

\label{sec:input_perturbation}
Particularly, we assume that a model that memorizes information should be able to produce the correct answer, even if the answer is not present in the input.
Thus, we ask the model a question about a specific key-value pair while removing all the clues related to the answer from both modalities (image and text).
Then, we consider that all correct answers are due to memorization.

We focus the experiments on generic keys of each provider e.g. name, email, tax number, etc. that are constant in all provider's documents and thus, more likely to be memorized. \Cref{table:provider_name} shows the results with \textit{provider\_name} as the key of interest. For more details of the test and results with different keys, please refer to the supplementary material.

\begin{table}[ht]
\caption{\textbf{DocVQA Performance of Non-Private Models on \memtest}. All information related to \textit{provider\_name} are removed from input.}
\label{table:provider_name}
\footnotesize
\centering
\begin{tabular}{|c|cc|cc|cc|cc|}
\hline
\multirow{3}{*}{Model} & \multicolumn{4}{c|}{} & \multicolumn{4}{c|}{$\conf\ge 0.9$} \\
\cline{2-9}
 & \multicolumn{2}{c|}{\redin} &  \multicolumn{2}{c|}{\redout} &   \multicolumn{2}{c|}{\redin}& \multicolumn{2}{c|}{\redout}\\

 & ACC & ANLS & ACC & ANLS & ACC & ANLS & ACC & ANLS\\
\cline{1-9}
VT5\textsubscript{0} & 0.08 & 1.60  & 0.09 & 0.68 & 0 & 1.16 & 0 & 0.72\\ 

VT5\textsubscript{C} & \textbf{\textcolor{red}{3.55}} & 11.64 & 0.17 & 5.05 & \textbf{\textcolor{red}{4.66}} & 14.53 & 0 & 6.60\\
\hline
\end{tabular}
\end{table}
\setlength{\tabcolsep}{1.4pt}

\Cref{table:provider_name} shows that while the \vtfc model fails to predict the name of providers in the \redout, it achieves around $3.5\%$ Accuracy ($4.6\%$ when highly confident) in the \redin even when no relevant information about the answer is available. Given its generative nature, this behavior suggests that the knowledge about these \redin providers is actually stored inside the model after fine-tuning. The memorization effect is further confirmed by the $0\%$ accuracy of model \vtfz in zero-shot setting, which confirms that this particular knowledge was not present in the model before fine-tuning.

\subsection{Attack Strategies}
Membership Inference Attack (MIA) aims to classify %
target samples as ones belonging to the training set (in) or not (out).
In this Section, we introduce three novel MIA that aim to systematically detect whether a particular provider's data has been used for training the model.

\subsubsection{Provider Membership Inference Attack (PMIA)} is a binary classification task that attempts to infer whether a specific provider $P$ contributes its data to the training set of a PFL-DocVQA target model. The attacker, owning a set of non-training documents from the provider $P$ %
can query the target model $h$ multiple times, each with one input tuple document, question and answer $(d_i, q_i, a_i)$ from the set of possible inputs $I_P$, %
and in return receive a set of outputs $O_P$. Then, the attack model $\mathcal{A}$ aggregates all of these outputs into a feature vector $f_P=\text{AGG}(O_P)$, where \text{AGG} is the feature aggregation function, and produces a binary prediction $\mathcal{A}(f_P;h)\in \{0;1\}$, indicating if $P$ comes from the training set.

Unlike prior works such as in~\cite{nasr2019comprehensive,Property,suri2022subject,shokri2017membership}, where attacks are evaluated directly on training examples, in this case we assume \textit{query data pertaining the provider $P$ can not be exactly the one used in training}%
, which we believe is more realistic in real-world scenarios.

\noindent

\subsubsection{Metric Selection}
\label{sec:metrics}
We base our attacks on a variety of selected per-example metrics
that reflect statistics of specific behaviours of the model, which can in turn be used to distinguish training data from non-training data. The combination of these metrics provides a good descriptor to identify membership.

We categorize our metrics into different groups $G_i$, noting that each metric is computed per-example:

\begin{itemize}
    \item $G_1 = (\textit{ACC}, \textit{NLS})$ are DocVQA utility metrics ANLS and accuracy. %
    In general, models tend to yield higher utility metrics for data sampled from their training set, 
    and this observation forms the basis for many state-of-the-art MIAs~\cite{shokri2017membership,suri2022subject}. It is also assumed that the model can output the loss value $L$ and prediction confidence $conf$, which constitute another metric group denoted as 
    $G_2 = (\mathit{L}, \mathit{\conf})$.
    \item $G_3 = (\Delta{L}, \Delta{\conf})$ are computed as the difference of the respective utility metrics of the model before and after fine-tuning%
    . Intuitively, we expect to see higher values for this group of metrics for providers who are part of the training data.
    \item $G_4 = (\textit{NLS}\textsubscript{mem}, \Delta{NLS}\textsubscript{mem})$ are two metrics inspired from \cref{sec:memorization}. $\textit{NLS}\textsubscript{mem}$ indicates the model's NLS score in the Memorization Test. %
    $\Delta{NLS}\textsubscript{mem}$ is designed in another test, where we first let the model generate \textit{the most likely output given no query}. We then measure how much the output changes if we remove it from the input and test the model again in the same setting. If it remains similar, this is a sign of memorization. %
\end{itemize}

\subsubsection{Proposed Approaches}
We propose two attack settings based on the knowledge the attacker has about the model and the training data.

\noindent
\textbf{Unsupervised Attack with Zero Knowledge (AZK).}
This attack corresponds to a \textit{zero-knowledge} setting, where only black-box access to the fine-tuned model and the ground-truth answer of each query are available to the adversary. The attacker has no information about the in set of providers $\Din$, neither about the target model architecture. 

In this case the feature vector is formed with \textit{ACC} and \textit{NLS}, since these are the only available metrics in this setting. We then run K-Means Algorithm on the features to find the two clusters of in/out 
providers. The cluster with higher average accuracy is considered the set of member providers. %

\noindent
\textbf{Supervised Attack with Partial Knowledge (APK).}
The second attack focuses on the \textit{partial-knowledge} setting, in which it is assumed that the attacker knows a small number of both member/non-member providers from the training set, which is widely considered in some previous work~\cite{nasr2019comprehensive,shokri2017membership,hu2022m,carlini2022membership}. Yet, the adversary is only aware of the identities of the providers, without actually having access to the documents utilized during the training process. 
We also assume that the model architecture is known by the adversary, since the training starts from a publicly available pre-trained model, while maintaining the black-box access to the pre-trained and fine-tuned models. Lastly, we consider that the model outputs are tuples containing also the loss and confidence value for each query, in addition to the accuracy and the NLS of the previous setting.

In this approach, we randomly sample $r$-percent of target providers, which is the set of providers in RED, where $r$ is the sampling rate, and obtain a training subset $\mathcal{P}_{\text{train}}$ with member/non-member classes equally distributed. We then evaluate the attack on the rest of providers $\mathcal{P}_{\text{test}}$. Since black-box access to both the pre-trained and fine-tuned model is given, we can make use of the 6 metrics from groups 1, 2 and 3, to enrich the information from one provider. Similarly to the first approach, we use the concatenation of aggregated metric values as the feature vector $f_p$ for each provider $P\in\mathcal{P}_{\text{train}}$ and train a Random Forest classifier to infer provider membership.

\textbf{Ensemble with Memorization.}
Given the memorization results illustrated in \cref{sec:memorization}, we combine our main approaches with the memorization features of $G_4$ into a Hard-Voting Ensemble to further boost the performance.
In particular, we have two separate classifers, denoted as $\text{CLS}_{1}$ and $\text{CLS}_{2}$. In $\text{CLS}_{1}$, we separate members from non-members by thresholding $\textit{NLS}\textsubscript{mem}$ at 0, while in $\text{CLS}_{2}$ we use K-Means to figure out the two clusters based on $\Delta\textsubscript{mem}{NLS}$.

\noindent

\section{Federated Learning and Privacy baselines}
\label{sec:methods}

In this Section, we introduce a baseline method that is based on fine-tuning the VT5 model using (DP) federated learning techniques in order to limit the leakage of sensitive information.

\subsection{Federated Learning}

To perform Federated Learning~\cite{McMahan2017fedavg,shokri2015privacy}, we apply the standard FedAvg~\cite{McMahan2017fedavg} strategy. For this, the model weights of the non-frozen layers are sent from the server to the selected clients at each federated learning round. The model is then trained in parallel by each of the clients, and then, the model update of the non-frozen layers is returned to the server, where the different updates are averaged to obtain an updated model state. This process is repeated for each federated learning round. We measure the efficiency of the communications as the total amount of information transmitted between the server and the clients in Gigabytes (GB).

\subsection{Differentially Private Federated Learning}\label{sec:dpfl}

\subsubsection{Differential Privacy (DP)}
\label{sec:dp}

(\Ebudget, $\delta$)-DP~\cite{dwork2006epsilondelta} applied to machine learning bounds how much the output distribution of a randomized training algorithm can differ on adjacent training datasets, i.e.\ training datasets differing in the contribution of a single entity. The allowed difference is measured by \EbudgetS$\geq 0$ (lower means more private) and $\delta \in [0,1]$ (commonly fixed at $10^{-5}$) that define the privacy budget.
We refer to Dwork and Roth~\cite{dwork2014algorithmic} for a thorough introduction to DP.

Applied to our baseline method, DP ensures that there is a high probability that a machine learning model trained under DP is similar whether or not the data of a particular provider has been used for training the model.
We seek to protect the privacy of providers that could be leaked through textual (company name, account number) or visual (logo, layout) information.
We realise this by using \emph{provider-level add/remove adjacency}, where adjacent training datasets can be obtained by adding or removing all documents from one provider. Prior work denotes this as group-level DP~\cite{marathe2022subject,Galli_2023}.

Our training algorithm uses DP stochastic optimisation \cite{dp-sgd-rajkumar-2012,dp-sgd-song-2013,abadi2016deep}. Its privacy guarantee is based on clipping the contribution from each unit of privacy (in our case each provider) and adding Gaussian noise. Random subsampling of clients and providers at each iteration provides further amplification of privacy. %

We consider an adversary who can access all intermediate states of the global model, %
inferring the training data (or some private information about them) focused on the participating clients' providers. Nevertheless, we consider a passive (i.e., honest-but-curious), that is, it does not modify the trained global model.

\subsubsection{DP Federated Learning algorithm}
Our private federated algorithm FL-PROVIDER-DP is shown in \cref{alg:group_fed_learn}. At every FL round the server randomly selects a set of clients. Each selected client runs a local instance of federated learning where each provider acts as the training data of a ``virtual client'' within the real client. The client randomly selects providers, clips the per-provider updates and the adds an appropriate amount of noise so that the update aggregated by the server is differentially private with respect to all providers in $\cup_k \mathbb{P}_k$ (thus the division of the per-client noise by the number of sampled clients $|\mathbb{K}|$). The noisy update of each client is normalized by a constant $M$ which is the minimum number of providers among all clients. Note when no clients are sampled in a FL round the server still needs to add noise.

\begin{algorithm}[H]
\small
		\caption{FL-PROVIDER-DP
  \label{alg:group_fed_learn}}
	\DontPrintSemicolon
	{\bf Server:}\;
	\Indp Initialize common model $\mbf{w}_0$\;
	\For {$t=1$ \KwTo $T_{cl}$}
	{
	    Select set $\mathbb{K}$ of clients randomly \;
		\For {each client $k$ \textrm{in} $\mathbb{K}$}
		{	
			$\mathbf{u}_{t}^{k}= \mathbf{Client}_k(\mbf{w}_{t-1}, |\mathbb{K}|)$\;
		}
		$\mbf{w}_{t} = \mbf{w}_{t-1} + \frac{1}{|\mathbb{K}|} \sum_{k \in \mathbb{K}} \mathbf{u}_{t}^{k}$\;
	}
	\KwOut{Global model $\mbf{w}_{T_{cl}}$}\;
	\Indm {\bf $\mathbf{Client}_{k}(\mbf{w}_{t-1}, |\mathbb{K}|)$:}\;
	\Indp
	$\mathbb{P}_k$ is a set of predefined disjoint providers in $D_k$\;

    Select $\mathbb{M} \subseteq \mathbb{P}_k$ randomly\;
    \For {each provider $P$ \textrm{in} $\mathbb{M}$}
    {	$\mbf{w}' = \mbf{w}_{t-1}$\;
        $\Delta \mbf{w}_t^P = \mathbf{AdamW}(P, \mbf{w}', \Tgd) - \mbf{w}_{t-1}$\;
        $\Delta \hat{\mbf{w}}_t^P = \Delta \mbf{w}_t^P / \max\left(1, \frac{||\Delta \mbf{w}_t^P||_2}{C}\right)$\;
    }
    $\mathbf{w}_{t}' = \mathbf{w}_{t-1} + 
    \frac{1}{M} \left( \sum_{i \in \mathbb{M}}  \Delta \hat{\mbf{w}}_t^i + \mathcal{N}\left(0, \frac{\sigma^2 C^2 \mathbf{I}}{|\mathbb{K}|}\right) \right ) $
	
    \KwOut{Model update $\left(\mbf{w}_{t}' - \mbf{w}_{t-1} \right)$}
\end{algorithm}

The privacy loss of FL-PROVIDER-DP follows the usual analysis of DP stochastic optimisation consisting of compositions of sub-sampled Gaussian mechanisms. The loss depends on the number of iterations $T_{cl}$, sub-sampling rate $q$ (both over clients and providers) and noise scale $\sigma$ \cite{koskela2020,koskela2021,Gopi2021prv}. Details in the supplement in \cref{sec:priv_analysis_appendix}.

\subsection{Centralized DP training}
The centralized DP training assumes that only one client exists and all providers are part of the dataset of this client. Our centralized DP training follows the same approach as the federated learning baseline with the exception that at every iteration $T_{cl}$ the only existing client is selected. 

\section{Experiments}

In our experiments we use five different variants of the VT5 DocVQA model:
\begin{itemize}
\item \textbf{\vtfz} \textit{(Zero-shot baseline)}:  This method is not fine-tuned on the new \datasetName dataset.

\item \textbf{\vtfc} \textit{(Centralized non-DP model)}: Fine-tuned for 10 epochs on the \datasetName dataset. It is used as an upper bound of question-answering performance with respect to the other variants.

\item \textbf{\vtfcdp} (\textit{Centralized DP model}): Fine-tuned for 10 iterations and in expectation 1000 providers are sampled at every iteration. 

\item \textbf{\vtffl} \textit{(Federated Learning non-DP model)}: Fine-tuned for 10 FL rounds, sampling $K=2$ clients at each FL round.

\item \textbf{\vtffldp} \textit{(Federated Learning DP model)}: Fine-tuned for 10 FL rounds, in expectation $K=2$ clients are sampled per FL round and no subsampling of providers is done.

\end{itemize}

\noindent For all variants, a learning rate of $2e^{-4}$ is used. For the DP variants the noise scale $\sigma$ is computed to reach the targeted privacy budget. %

\subsection{Model Performance} \label{subsec:docvqa_performance}
\begin{figure}[t]
    \centering
    \includegraphics[width=\textwidth]{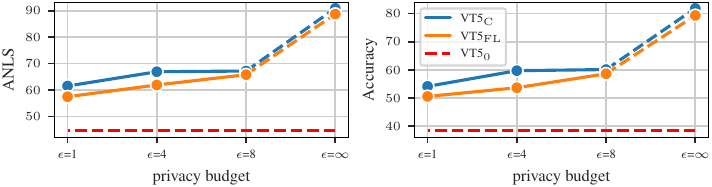}
    \caption{Question answering performance of the base method VT5. The DP models are trained with $\delta=10^{-5}$.}
    \label{fig:docvqa_results_plot}
\end{figure}

The question-answering results of the VT5 variants is shown in \cref{fig:docvqa_results_plot}. As expected, the \vtfc performs the best. \vtffl performance is close to the centralized version despite training the same amount of iterations and seeing only $20\%$ of the data at each iteration. In contrast, the private \vtfcdp and \vtffldp performance degrades due to the introduced noise and clipping, which increases as the privacy budget \EbudgetS becomes more limited. Tabular results are shown in \cref{tab:results_utility_blue_extended} the supplementary material.%

\subsection{Attack Performance}

\begin{figure}[t]
    \centering
    \includegraphics[width=\linewidth]{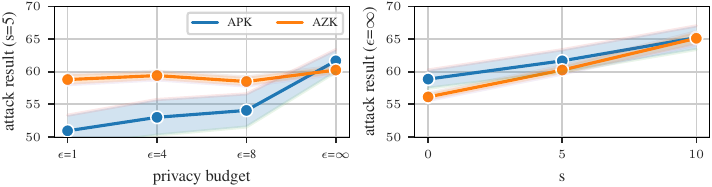}
    \caption{The left panel displays attack performance as a function of privacy budget for $s=5$ whereas the right panel displays the performance as a function of $s$ for the non-private model. AZK/APK denotes our \textit{zero-knowledge}/\textit{partial-knowledge} setting. Results are reported with the standard deviation over 5 random seeds.}
    \label{fig:pmia_results}
\end{figure}

Our attack experiments are conducted on the RED data, which contains $8,100$ Questions from 307 $\redin$ / 353 $\redout$ Providers (member/non member). 
The attack's performance is presented in terms of the Attack Accuracy metric. We evaluate our proposed attacks on multiple subsets $T_s$, with their size controlled by $s$, in such a way that $T_s$ only includes those providers with a minimum of $s$ defined questions. %
Increasing $s$ guarantees a set of informative providers with sufficient statistics while keeping lower $s$ requires the attack model to be robust to outliers. %
In \cref{fig:pmia_results} we plot the performance of our proposed attacks while provide the detailed results in \cref{tab:pmia_results} in supplementary material.

\subsubsection{Non-private model}
\label{sec:attack_results}

For the centralized non-private model, we can see that both approaches consistently perform better as $s$ increases, suggesting that a certain number of queries is expected to characterize a provider membership (\cref{fig:pmia_results} right panel). Under the simple assumption of \textit{zero-knowledge}, AZK can surpass APK on all three benchmarks (\cref{tab:pmia_results} row 1 and 3) with the same set of features, and even perform on par with fewer features on $T_{10}$ (\cref{tab:pmia_results} row 1 and 4). This implies $G_1$ metrics can effectively differentiate between members and non-members with enough queries. Still, AZK fails if less information are available, as simple K-Means algorithm is quite sensitive to outliers. 

When $G_2$ and $G_3$ features are added, APK clearly outperforms the unsupervised approach across all the evaluation sets (\cref{tab:pmia_results} rows 1 and 4), particularly in the high-data regime by a margin of 3.5\% Accuracy in $T_{0}$ and 1.4\% Accuracy in $T_{5}$, demonstrating the benefits of training a model with \textit{partial-knowledge}, combined with enriched feature vector. 

Memorization metrics $G_4$ are useful, especially effective when combined with $\text{CLS}_{1}$ and $\text{CLS}_{2}$ into Ensemble, as it improves attack's performance in every subsets for both approaches, with the best Accuracy at almost 67\% in $T_{10}$. This is not always the case if it is used as training features to the attack model (\cref{tab:pmia_results} rows 2 and 5). We hypothesize this observation as Memorization Test can supply a clear sign of memorized providers which other features can not provide. However, since this test falls into low-recall regime, as shown in \cref{table:provider_name} where the model fails most of the time even with training providers, using these features to train rather introduces noise to the model.

\subsubsection{Private models}

We take the best performing attack configurations presented in the previous section and we apply them to the centralized DP model described in \cref{sec:methods} in order to assess the ability of the proposed DP framework to reduce privacy leakage. %

Results in \cref{fig:pmia_results,tab:pmia_results} show that for both types of attacks, AZK and APK, the accuracy of the attack in the centralized-DP model is significantly lower than the accuracy in the non-private model for the same configurations. This reduction in the accuracy of the attack is more relevant in the case of the APK method that has a greater potential of leaking private information. 
These results confirm that the proposed DP baseline can effectively mitigate privacy leakage for DocVQA.

\section{Conclusions}
In this paper, we have explored for the first time privacy issues related to  DocVQA %
by proposing a large scale DocVQA dataset especially designed for preserving the identity of the providers used to train the model. We have shown that state of the art generative multi-modal models exhibit memorization and how this can be used to attack the model and reveal private information about the providers. As a solution to this privacy leakage we have proposed to use Federated Learning and Differential Privacy in a baseline framework that guarantees a higher level of privacy at the cost of reducing the utility of the model. 

{%
\section*{Acknowledgements}

This work has been funded by the European Lighthouse on Safe and Secure AI (ELSA) from the European Union’s Horizon Europe programme under grant agreement No 101070617.
RT, KN, MAS, LK, EV and DK have been supported by the Consolidated Research Group 2017-SGR-1783 from the Research and University Department of the Catalan Government, the projects PDC2021-121512-I00, PID2020-116298GB-I00 and PLEC2021-007850 funded by the European Union NextGeneration EU/PRTR and MCIN/AEI/10.13039/501100011033. %
MT, JJ and AH have been supported by the Research Council of Finland (Flagship programme: Finnish Center for Artificial Intelligence, FCAI; as well as grants 356499 and 359111) and the Strategic Research Council at the Research Council of Finland (Grant 358247). %
Part of this work has been performed using resources provided by the CSC – IT Center for Science, Finland, and the Finnish Computing Competence Infrastructure (FCCI).
Views and opinions expressed are however those of the author(s) only and do not necessarily reflect those of the European Union or the European Commission. Neither the European Union nor the granting authorities can be held responsible for them. 
}

\newpage

\bibliographystyle{splncs04}
\bibliography{main}

\newpage
\appendix
\setcounter{figure}{0}
\renewcommand\thefigure{A.\arabic{figure}}
\setcounter{table}{0}
\renewcommand{\thetable}{A\arabic{table}}
\setcounter{equation}{0}
\renewcommand{\theequation}{A\arabic{equation}}

\section{Dataset}\label{sec:supp_dataset}

\begin{figure}[h]
    \centering
    \begin{tabular}{|c|c|c|}
        \hline
        \includegraphics[width=0.3\columnwidth, height=2.9cm]{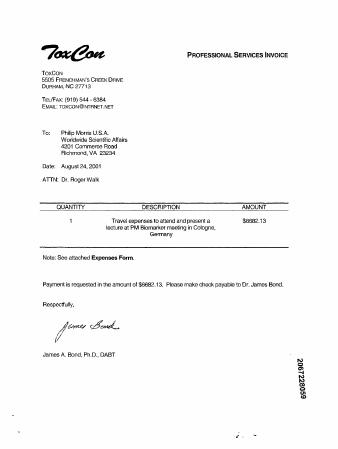} & \includegraphics[width=0.3\columnwidth, height=2.9cm]{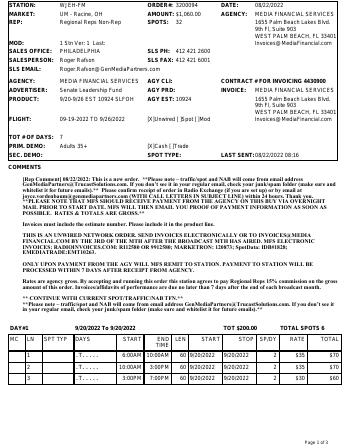} &
         \includegraphics[width=0.3\columnwidth, height=2.9cm]{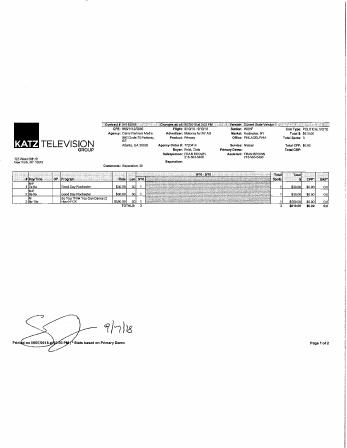} 
         \\\hline
         \includegraphics[width=0.3\columnwidth, height=1.5cm]{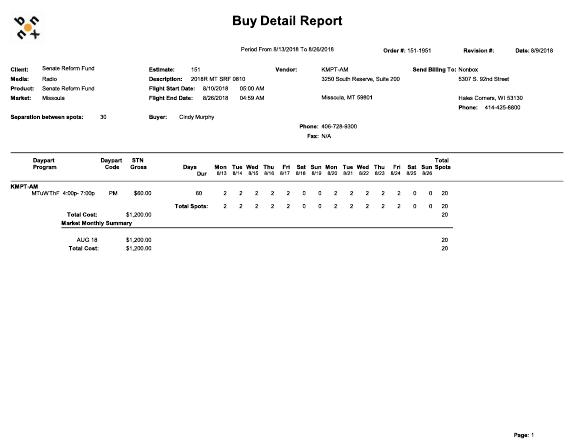} & \includegraphics[width=0.3\columnwidth, height=1.5cm]{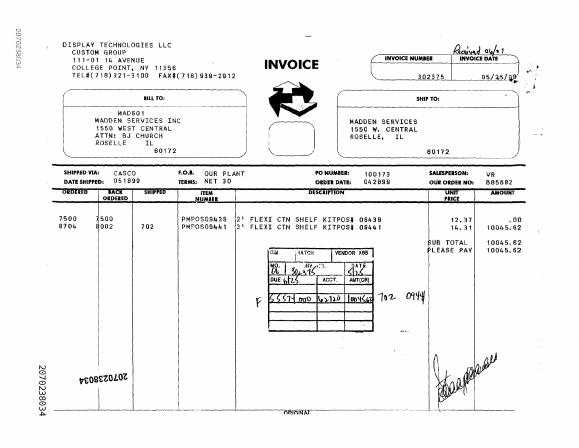} &
         \includegraphics[width=0.3\columnwidth, height=1.5cm]{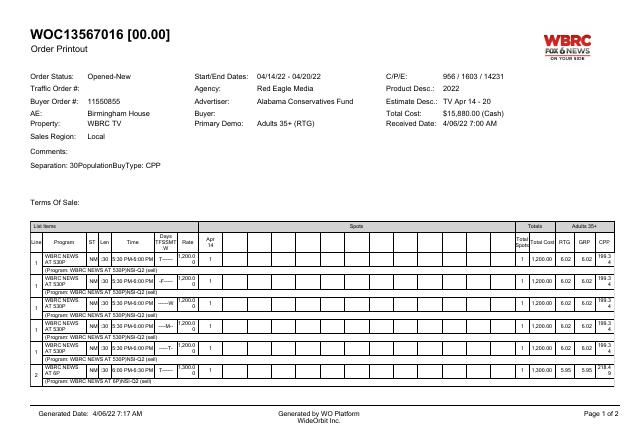} 
         \\\hline
         \includegraphics[width=0.3\columnwidth, height=2.9cm]{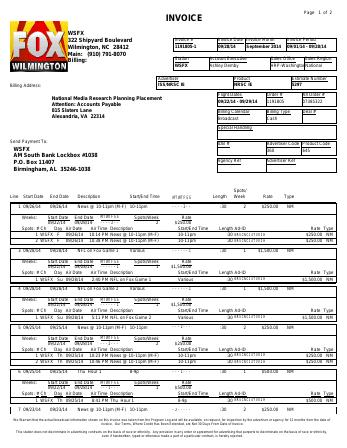} & \includegraphics[width=0.3\columnwidth, height=2.9cm]{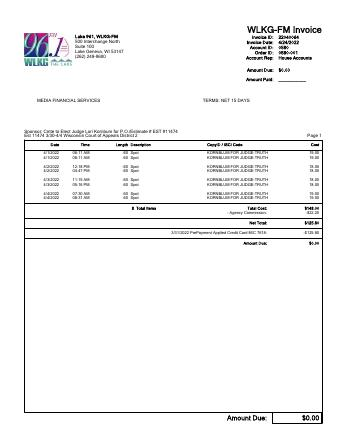} &
         \includegraphics[width=0.3\columnwidth, height=2.9cm]{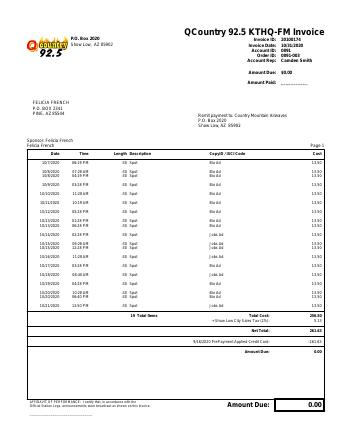} 
         \\\hline
    \end{tabular}
    \caption{Examples of different invoice document images of from \datasetName.}
    \label{fig:docs_examples}
\end{figure}

\begin{figure}[H]
    \centering
    \begin{tabularx}{1.1\linewidth}{lp{0.13\textwidth}p{0.50\textwidth}}
        \multirow{ 6}{*}{\includegraphics[width=0.5\columnwidth]{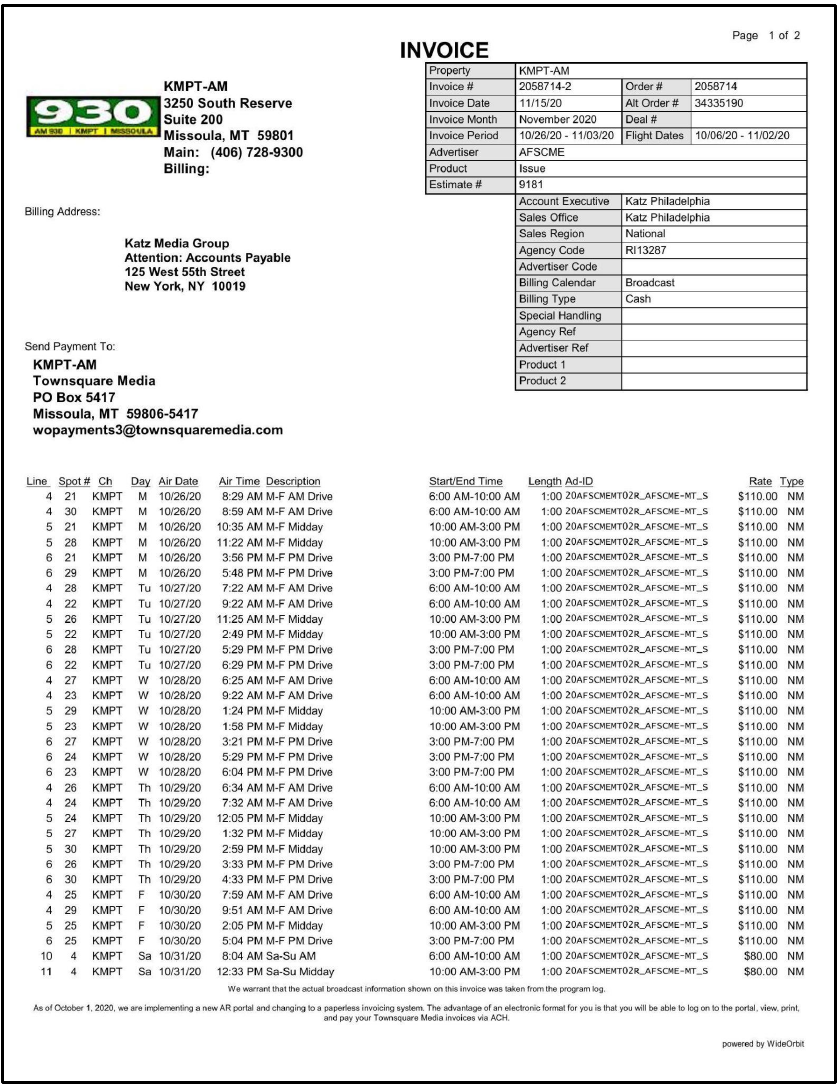}} & \textbf{Question:} & Can you provide the code associated with the agency mentioned in the document? \\
        &   \textbf{Answer:} & RI13287 \\
        & & \\
        & \textbf{Question:} & What month is attributed to the invoice? \\
        &   \textbf{Answer:} & November 2020 \\
        & & \\
        & \textbf{Question:} & Can you inform me of the vendor's name? \\
        &   \textbf{Answer:} & kmpt-am \\
        & & \\
        & \textbf{Question:} & Can you specify the date the invoice was printed? \\
        & \textbf{Answer:} & 10/17/18 \\
    \end{tabularx}
    \vspace{2.5cm}
    \caption{Examples of questions and answers on an invoice document within \datasetName.}
    \label{fig:sample-data}
\end{figure}

 The \datasetName dataset is built from invoice documents that belong to different providers and clients, some examples of the invoice documents are shown in \cref{fig:docs_examples}.  \datasetName is composed of the document images, the OCR tokens and a set of questions/answers for each page. An example of these questions and answers is provided in \cref{fig:sample-data}.
 
\begin{figure}[h]
    \centering
    \begin{tabular}{|c|c|c|}
        \hline
         \includegraphics[width=0.3\columnwidth]{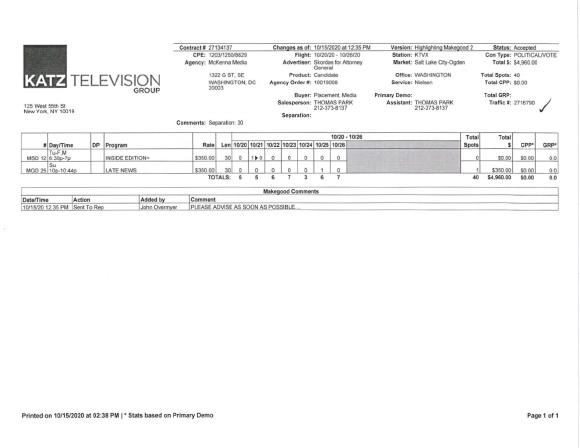} & \includegraphics[width=0.3\columnwidth]{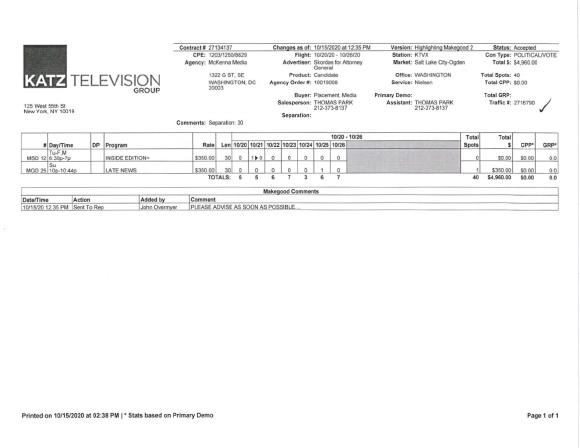} &
         \includegraphics[width=0.3\columnwidth]{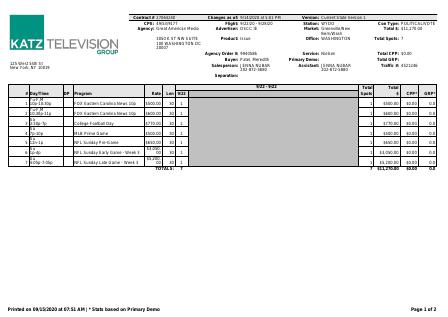} 
         \\\hline
         \includegraphics[width=0.3\columnwidth]{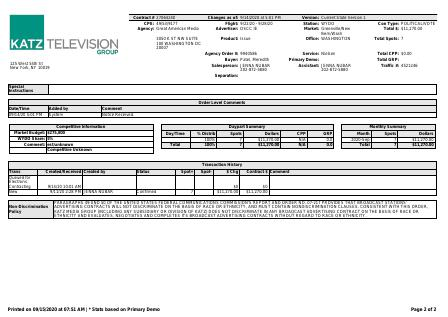} & \includegraphics[width=0.3\columnwidth]{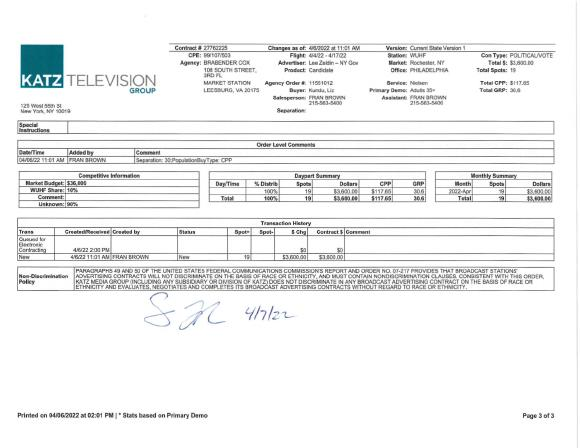} &
         \includegraphics[width=0.3\columnwidth]{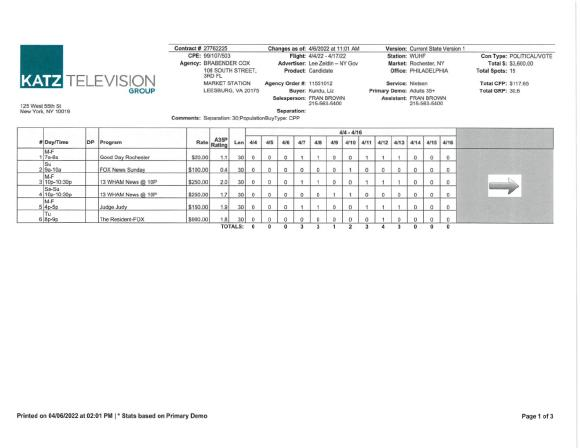} 
        \\\hline
         \includegraphics[width=0.3\columnwidth]{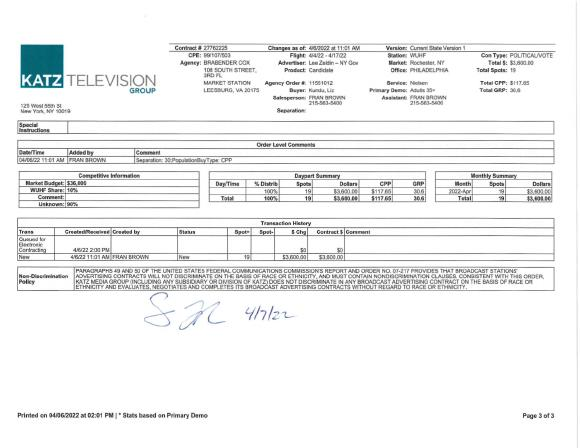} & \includegraphics[width=0.3\columnwidth]{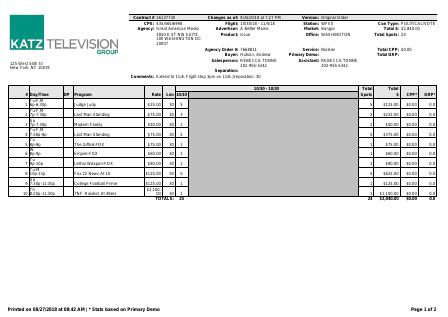} &
         \includegraphics[width=0.3\columnwidth]{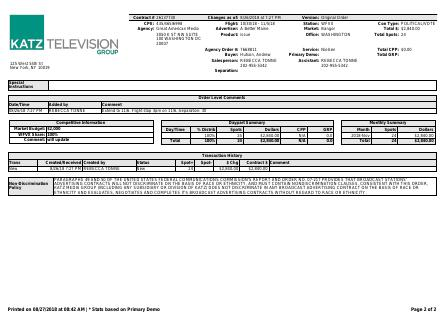} 
         \\\hline
    \end{tabular}
   
    \caption{A visualization of different invoice documents from the same provider. The documents have similar layouts and use the same logo.}
    \label{fig:provider_docs}
\end{figure}

\noindent The dataset is divided into BLUE and RED sub-datasets, as described in \cref{sec:dataset} and shown in \cref{fig:data_splits}. To create these subsets, we first cluster the documents by the provider ID to define the $\mathcal{D}_{in}$ and $\mathcal{D}_{out}$ distributions, and split the data accordingly. In \cref{tab:pfl_docvqa_stats} we show the number of documents, pages and questions/answers per client/subset in the BLUE and RED data respectively.
Moreover, the documents of the same provider usually share visual and textual features, such as the invoices showed in \cref{fig:provider_docs}, which belong to KATZ TELEVISION.

\newpage

\begin{table}[H]
    \centering
    \scalebox{0.84}{
    \begin{tabular}{|c|c|c|c|c|c|c|}
    \hline
    \multirow{2}{*}{Dataset}                & \multirow{2}{*}{Split}                   & \vtop{\hbox{\strut Client}\hbox{\strut(Subset)}} & \multirow{2}{*}{Provider} & \multirow{2}{*}{Document} & \multirow{2}{*}{Page} & \vtop{\hbox{\strut Question/}\hbox{\strut Answer}} \\ \hline
    \multirow{13}{*}{BLUE} & \multirow{10}{*}{Train} & 0             & 400      & 2224         & 5930      & 19465           \\
                           &                         & 1             & 418      & 2382         &  6694     & 22229           \\
                           &                         & 2             & 404      & 2296         &  6667     & 21673           \\
                           &                         & 3             & 414      & 2358         &  6751     & 22148           \\
                           &                         & 4             & 429      & 4543         &  12071    & 32472           \\
                           &                         & 5             & 423      & 2378         &  6984     & 22361           \\
                           &                         & 6             & 423      & 2700         &  7406     & 23801           \\
                           &                         & 7             & 416      & 1951         &  5617     & 18462           \\
                           &                         & 8             & 401      & 1932         &  5421     & 17868           \\
                           &                         & 9             & 421      & 2136         &  6353     & 20840           \\ \cline{2-7} 
                           & \multirow{2}{*}{Valid}                   & Positive             & 1268     & 2561         &  --     & --           \\
                           &                    & Negative& 963 &975&--&-- \\
                           \cline{2-7} 
                           & \multirow{2}{*}{Test}   & Positive      & 1390     & 2875         &  8088     & 25603           \\
                           &                         & Negative      & 977      & 1912         &  5375     & 17988           \\ \hline
    \multirow{2}{*}{RED}   & \multirow{2}{*}{Test}   & Positive      & 307      & 425         &  1361     & 4205            \\
                           &                         & Negative      & 353      & 425         &  1228     & 3895            \\ \hline 
    \end{tabular}
    }
    \vspace{0.3cm}
    \caption{\textbf{Statistics on PFL-DocVQA Dataset} in terms of number of Providers/Documents/Pages/Question-Answers. The notion of \textit{client} is only applied to BLUE Train data, while BLUE/RED Test is divided into two subsets: Positive and Negative that are from $\mathcal{D}_{in}$ and $\mathcal{D}_{out}$, respectively.} 
    \label{tab:pfl_docvqa_stats}
\end{table}

\newpage

\begin{figure}[H]
    \centering
    \begin{tabular}{c}
        \includegraphics[width=0.61\columnwidth]{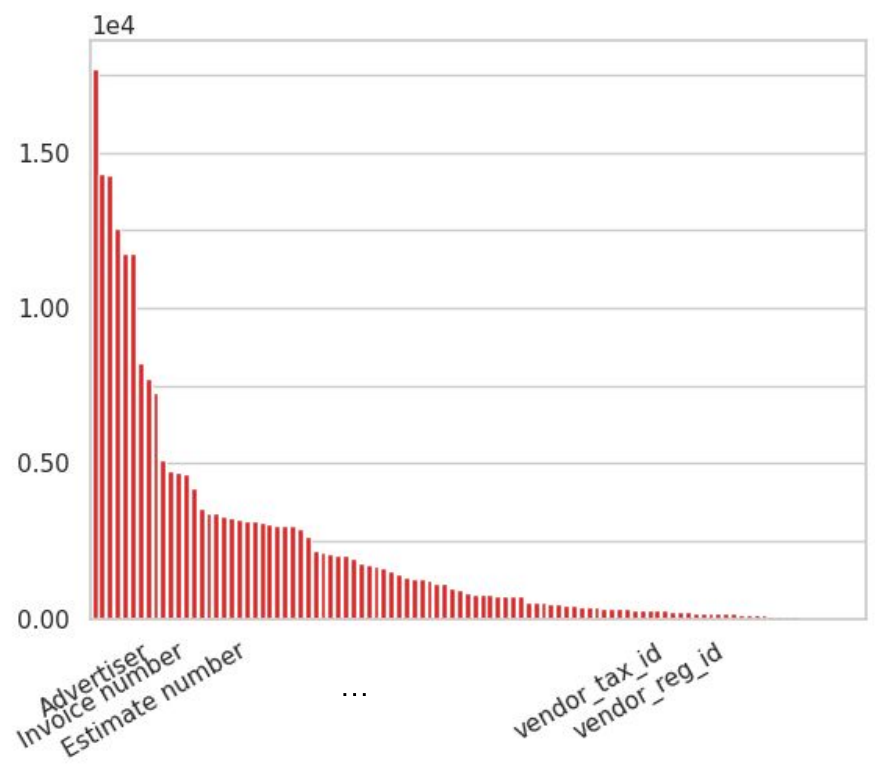}  
    \end{tabular}
    \caption{\textbf{Histogram of number of questions per key} in the training data.}
    \label{fig:question_distribution}
\end{figure}

\vspace{-0.5cm}
\noindent Moreover, we provide  an overview of the number of questions asked for each key in the training set in \cref{fig:question_distribution}.

\newpage

\section{Visual T5 Model}\label{sec:appendix_vt5}

\begin{figure*}[ht]
    \centering
    \includegraphics[width=0.96\linewidth]{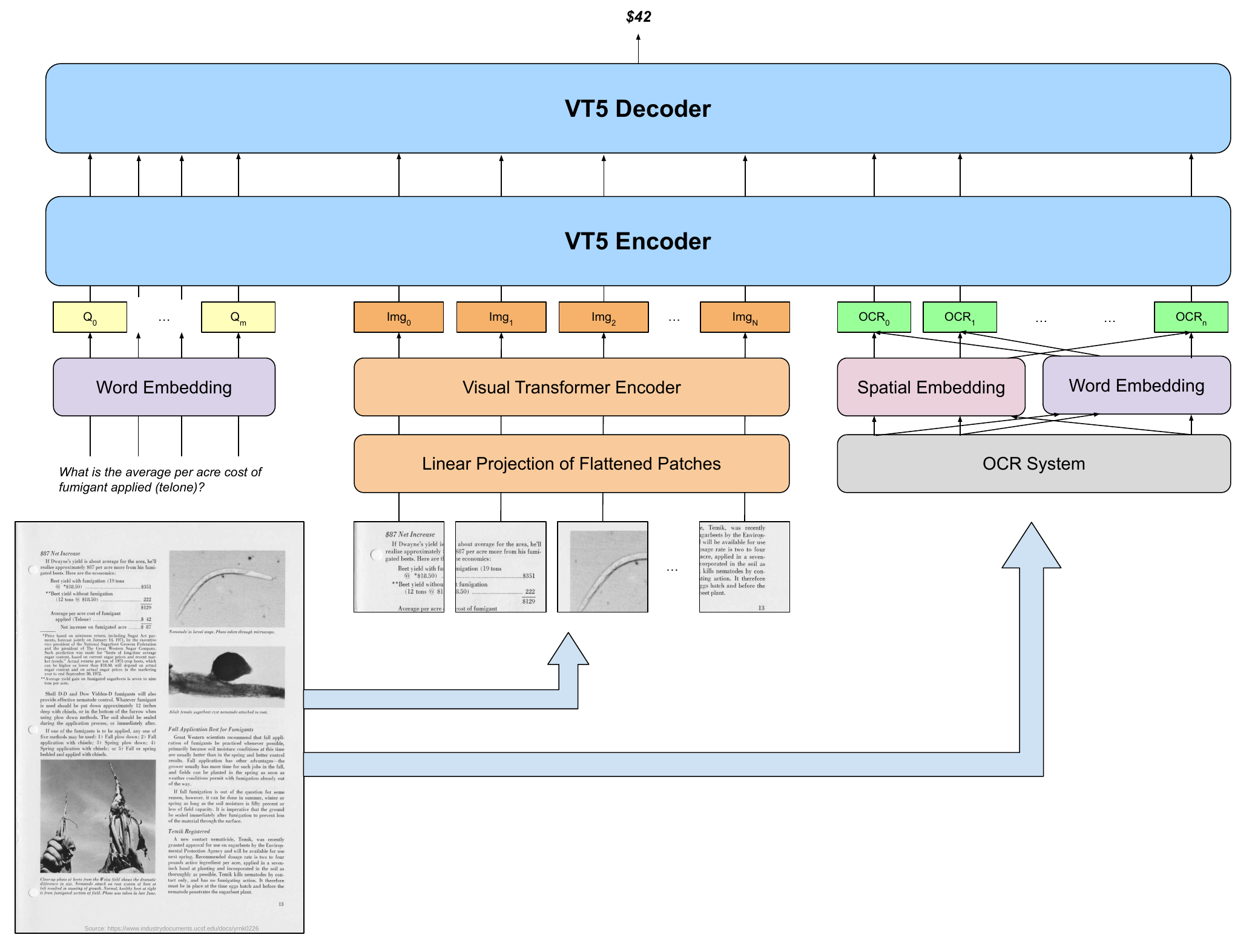}
    \caption{VT5 detailed method architecture.}
    \label{fig:vt5_detailed}
\end{figure*}

In this section, we provide a further detailed description of the employed VT5 base method illustrated in \cref{fig:vt5_detailed}.
The question and OCR words are first tokenized into subword tokens and embedded into a learned semantic contiguous representation from T5~\cite{raffel2020exploring}, which captures the contextual information and semantics of the token. Moreover, VT5 utilizes a spatial embedding to represent the spatial information from the bounding box using a lookup table for continuous encoding of one-hot vectors. The different $x_{0}, y_{0}, x_{1}, y_{1}$ are embedded into $2$ continuous learned embeddings for the horizontal and vertical coordinates. Then, the four embedded coordinates are aggregated and summed to the semantic representation. In parallel, the page image is divided into non-overlapping patches, which are projected and fed to the Visual Transformer DiT~\cite{li2022dit}. Then, all the features are fed into the T5 backbone self-attention layers to finally generate the answer in an autoregressive way.

\newpage

\section{Model Performance}

In \cref{tab:results_utility_blue_extended} we detail the results obtained in \cref{subsec:docvqa_performance} and plotted in \cref{fig:docvqa_results_plot}. %

\begin{table}[H]
\centering
\begin{tabularx}{0.75\columnwidth}{p{0.18\hsize} P{0.10\hsize} P{0.12\hsize} P{0.12\hsize} c}
\toprule
\multirow{3}{*}{Method} & \multirow{3}{*}{$\varepsilon$}  & \multicolumn{2}{c}{Question}  & \multirow{2}{*}{Communication} \\
                        &                                 & \multicolumn{2}{c}{Answering} & \multirow{2}{*}{(GB)}          \\
                        &                    & ANLS                & Acc.               &   \\
\midrule
\vtfz        & -        & 44.58    & 38.43     & -      \\
\vtfc        & $\infty$ & 90.91    & 81.80     & -      \\
\vtfcdp      & 8        & 67.19    & 60.12     & -      \\
\vtfcdp      & 4        & 66.92    & 59.72     & -      \\
\vtfcdp      & 1        & 61.52    & 54.16     & -      \\
\vtffl       & $\infty$ & 88.73    & 79.30     & 44.66  \\
\vtffldp     & 8        & 65.81    & 58.62     & 44.66  \\
\vtffldp     & 4        & 61.91    & 53.68     & 44.66  \\
\vtffldp     & 1        & 57.47    & 50.60     & 44.66  \\
\bottomrule
\end{tabularx}

\vspace{0.5cm}
\caption{Question answering performance and communication cost of the base method VT5 on the different set-ups. DP training is done with $\delta=10^{-5})$.%
}
\label{tab:results_utility_blue_extended}
\end{table}

\section{Attacks}\label{sec:supp_pmia}
\subsection{\memtest}\label{sec:supp_memorization_test}
\subsubsection{Experimental Setup}\label{sec:memorization_test_setup}
\begin{figure}[H]
    \centering
    \includegraphics[width=0.8\linewidth]{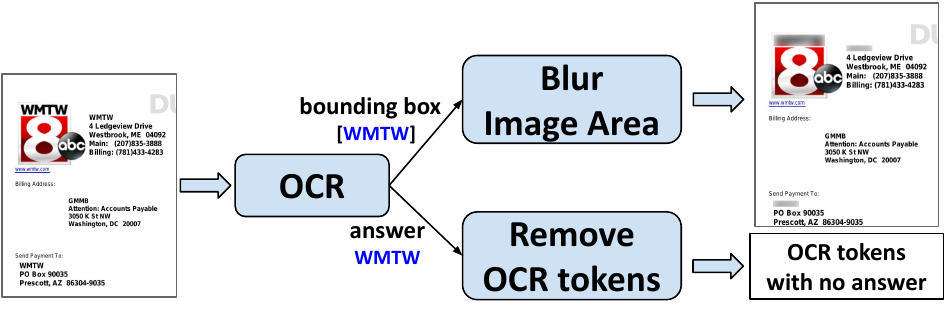}
     \\ ~ \\
     \begin{tabular}{l}
        \textbf{Question:} What name is associated with the vendor? \\
        \textbf{Answer:} \gt{WMTW} \\
     \end{tabular}
    \caption{\textbf{Our procedure to hide information in our \memtest} described in \cref{sec:memorization_test_setup} with an example.}
    \label{fig:memorization_test_procedure}
\end{figure}

\begin{figure}[H]
    \centering
    \setlength{\tabcolsep}{5pt}
    \scriptsize
    \begin{tabularx}{1.2\linewidth}{p{0.33\linewidth} p{0.33\linewidth} p{0.33\linewidth}}
        \frame{\includegraphics[width=\linewidth]{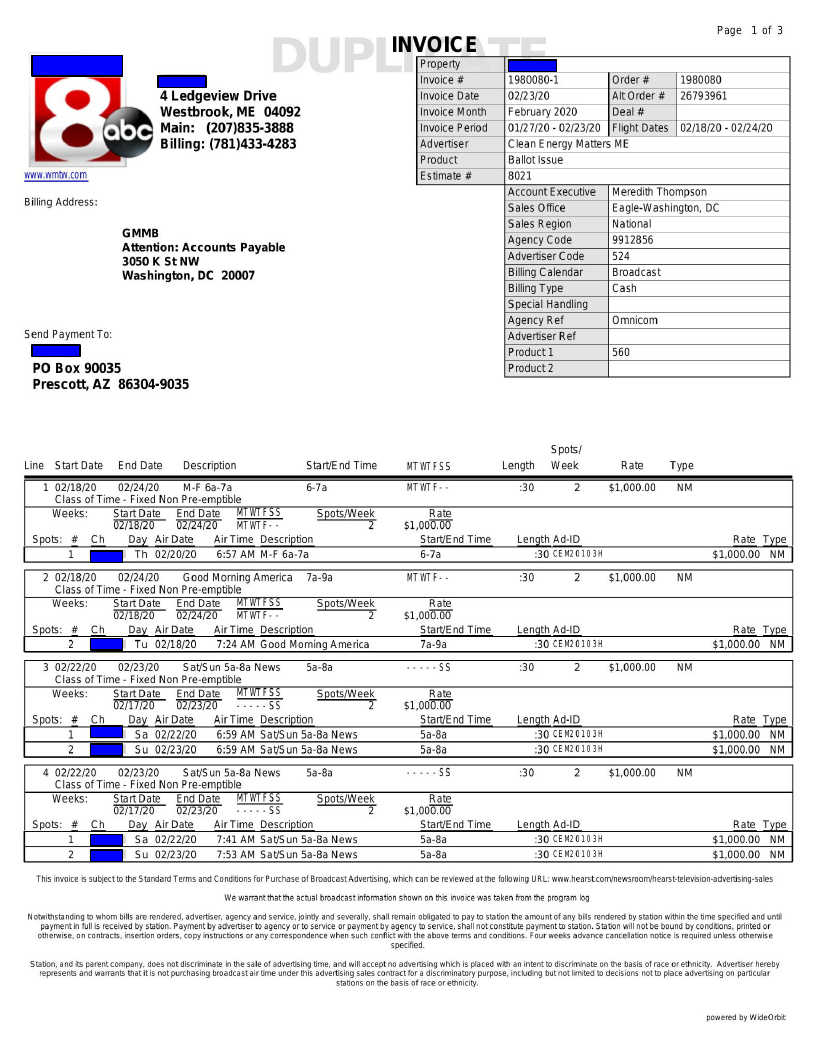}} & \frame{\includegraphics[width=1.015\linewidth]{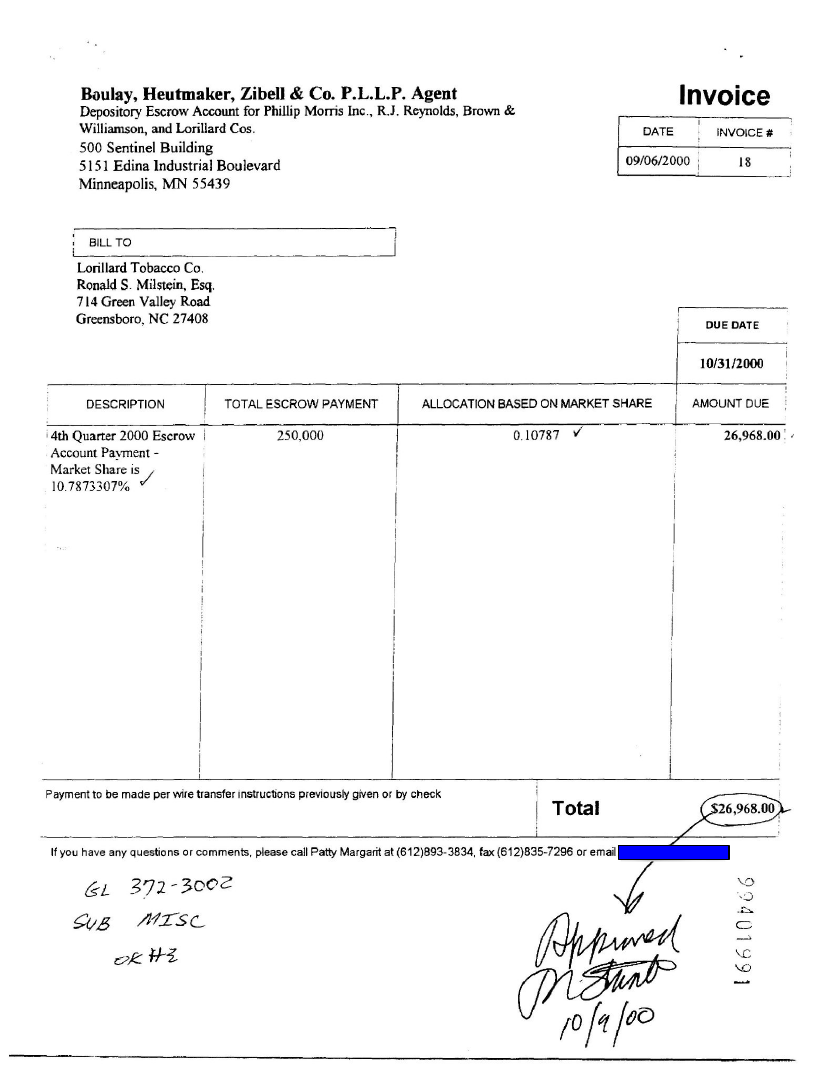}} & 
        \frame{\includegraphics[width=0.94\linewidth]{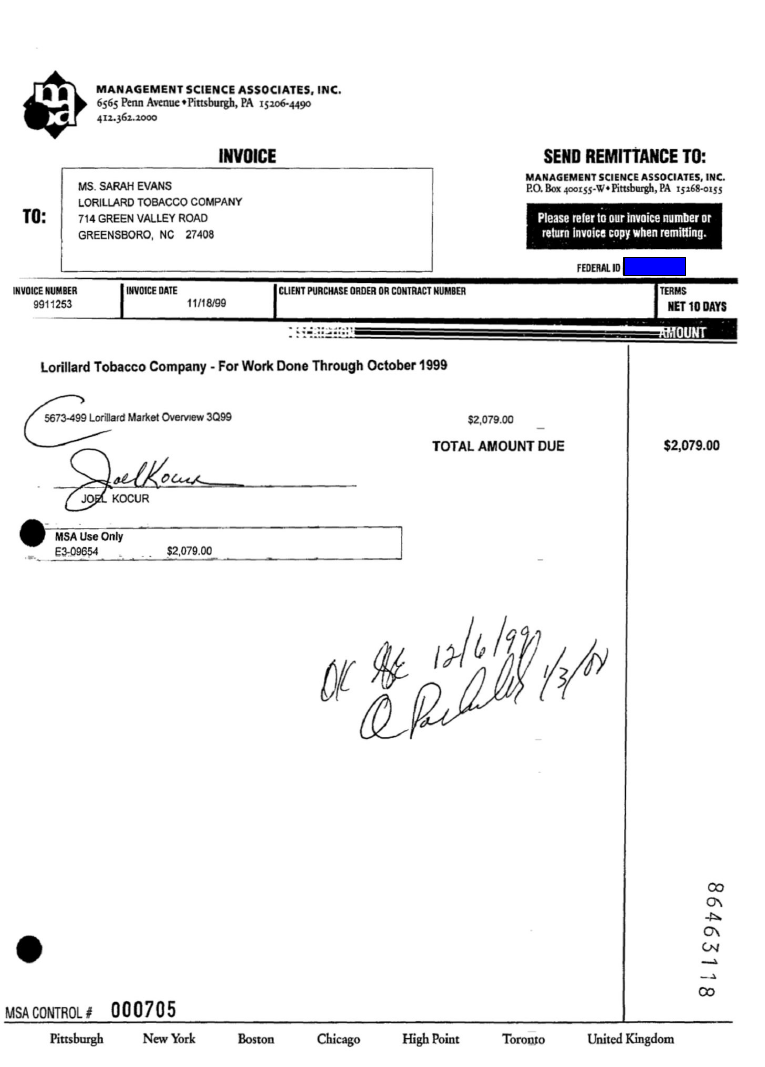}} \\
        
        \textbf{Key: Provider Name}  &
        \textbf{Key: Provider Email} &
        \textbf{Key: Provider Tax Number} \\
        
        \textbf{Quest.:} What is the provider of this document? &
        \textbf{Quest.:} Could you share the vendor's email? &
        \textbf{Question:} Can you let me know the vendor's tax ID?  \\

        \textbf{Answer:} \textcolor{red}{wmtw} &
        \textbf{Answer:} \textcolor{red}{pmargarit@bhz.com} &
        \textbf{Answer:} \textcolor{red}{25-1126415} \\
     \end{tabularx}
    \caption{\textbf{Examples of \memtest samples with information removed.} The blurred image area is colored \textbf{\textcolor{blue}{blue}} for better visualization.} 
    \label{fig:memorization_test_examples}
\end{figure}

\cref{fig:memorization_test_procedure} illustrates the procedure that is applied in each \memtest. To begin, we first select a \textit{key} that corresponds to a specific information of a provider, such as \textit{name, email, tax number etc.} 
Given the selected \textit{key}, we create a new test set which contains all RED documents, in each document we ask one question w.r.t the \textit{key}. The question is sampled from the set of pre-defined templates for each key in the training phase, while the answer is the ground-truth information.
Next, we remove all the clues related to the answer from both visual and textual input. For the visual part, we use Gaussian Blurring with the radius of Gaussian kernel set to 20. For the textual part, we discard all tokens in the OCR that are identified as exact or fuzzy matches to the answer string.
Note, the size of each \memtest might vary, depending on the availability of the ground-truth answer in each document. 
For instance, while the provider's name is present in most of the invoice documents, 
not all of them specify the provider email or address, making these documents unavailable for testing.
In addition, we skip all documents for which our code fails to process, maintaining the quality of the test.
Finally, we evaluate the model on this new evaluation set in terms of DocVQA utility metrics, where high scores indicate clear memorization behaviors. In \cref{fig:memorization_test_examples} we show some examples of the \memtest with the information that is asked removed.

\newpage
\subsubsection{More on Key: Provider Name}
We further analyze the results of \vtfc on the \memtest with \textit{provider\_name} reported in \cref{tab:supp_provider_name}, especially with the unexpected $0.17\%$ in Accuracy (and $5.05\%$ ANLS) on RED Negative, while it was expected to be 0.
By examining the predicted answers, it is suspected that this performance comes from residual information from the OCR. The model might have learned to make use of other OCR tokens that still contain information about the provider (e.g. our code fails to remove `www.kcci.com', which is the website for provider `kcci'). Nevertheless, near $0\%$ in Accuracy in this test means the model struggles to recover \textbf{the full answer}. 
Overall, we discover 5/340 positive providers with perfect Accuracy for all of their documents. %
\vspace{-0.5cm}
\begin{table}[H]
\footnotesize
\centering
\begin{tabular}{|l|cc|cc|}
\hline
\multicolumn{1}{|c|}{\multirow{3}{*}{Model}} & \multicolumn{2}{c|}{Red Positive}                 & \multicolumn{2}{c|}{Red Negative}                 \\ \cline{2-5} 
\multicolumn{1}{|c|}{}                       & \multicolumn{2}{c|}{$\text{N}_{\text{mem}}$=1333} & \multicolumn{2}{c|}{$\text{N}_{\text{mem}}$=1178} \\ \cline{2-5} 
\multicolumn{1}{|c|}{}                       & \multicolumn{1}{c|}{ACC}          & ANLS          & \multicolumn{1}{c|}{ACC}           & ANLS         \\ \hline
\vtfc                                         & \multicolumn{1}{c|}{\textbf{\textcolor{red}{3.55}}}         & 11.64         & \multicolumn{1}{c|}{\textbf{\textcolor{red}{0.17}}}          & 5.05         \\ \hline
\vtfcdp ($\varepsilon$=8)                     & \multicolumn{1}{c|}{0.03}         & 2.4           & \multicolumn{1}{c|}{0.03}          & 0.66         \\ \hline
\vtffldp ($\varepsilon$=8)                    & \multicolumn{1}{c|}{0.15}         & 2.86          & \multicolumn{1}{c|}{0.08}          & 1.72         \\ \hline
\end{tabular}
\vspace{0.25cm}
\caption{\textbf{\memtest with key \textit{provider\_name}}. $\text{N}_{\text{mem}}$ denotes number of testing documents. Compared to the Non-Private Model \vtfc, memorization on positive providers are much reduced with Private Models.}
\label{tab:supp_provider_name}
\end{table}

\subsubsection{Key: Provider Email}
To perform this test, we first construct the evaluation set as explained in \cref{sec:memorization_test_setup}. For any RED Positive provider, if there is any training question about the key posed on provider's documents, we reuse the training answer as the ground-truth answer for the test. Since this process cannot be applied to the RED Negative providers, we only use documents that have questions about the key found in RED testing examples. Similar to the pattern observed with the provider names, we identify a certain degree of memorization regarding positive providers' emails inside \vtfc, with $13\%$ of Accuracy in the test, which is highly mitigated after the introduction of DP in our training process. 
\vspace{-0.5cm}
\begin{table}[h]
\footnotesize
\centering
\begin{tabular}{|l|cc|cc|}
\hline
\multicolumn{1}{|c|}{\multirow{3}{*}{Model}} & \multicolumn{2}{c|}{Red Positive}             & \multicolumn{2}{c|}{Red Negative}             \\ \cline{2-5} 
\multicolumn{1}{|c|}{}                       & \multicolumn{2}{c|}{$\text{N}_{\text{mem}}$=133} & \multicolumn{2}{c|}{$\text{N}_{\text{mem}}$=8} \\ \cline{2-5} 
\multicolumn{1}{|c|}{}                       & \multicolumn{1}{c|}{ACC}        & ANLS        & \multicolumn{1}{c|}{ACC}        & ANLS        \\ \hline
\vtfc                                         & \multicolumn{1}{c|}{\textbf{\textcolor{red}{13.09}}}           &      32.65       & \multicolumn{1}{c|}{0}           &      39.91       \\ \hline
\vtfcdp ($\varepsilon$=8)                     & \multicolumn{1}{c|}{0}           &      1.39       & \multicolumn{1}{c|}{0}           &       0      \\ \hline
\vtffldp ($\varepsilon$=8)                    & \multicolumn{1}{c|}{0}           &      1.33       & \multicolumn{1}{c|}{0}           &       0      \\ \hline
\end{tabular}
\vspace{0.25cm}
\caption{\textbf{\memtest with key \textit{provider\_email}}. Memorization about this information is also observed with Non-Private Models.} 
\label{tab:supp_provider_email}
\end{table}

\newpage

\subsubsection{Key: Provider Tax Number, Registration ID}

Our experiments in \cref{sec:memorization_test_setup} uncover memorization behaviors within the target model, but the test is only verifiable in case the information is actually present in the testing documents. In this section, we reveal another weakness of this model, where 
the model can provide sensitive information to specific queries even if the information does not exist in the document. 

The focus of this test is on \textit{Tax Number/Registration ID} as the key information for two reasons: First, these details uniquely identify a specific entity. Second, these keys are considered as \textit{under-trained} as shown in \cref{fig:question_distribution}, which indicates the model has not learned sufficiently to answer these types of questions due to limited training data. As a result, any not-requested disclosure of such information is likely attributed to memorization.

We carry out the test on RED Negative providers, where we sample at maximum 5 different documents  per provider. %
For each document, we ask the target model \vtfc 2 questions about Tax Number/Registration ID. Note that, we do \textit{not hide} information before inference, but discard documents where the answer is explicitly present, as we aim to produce memorized answers, instead of actual answers from the document.
We list the top-15 most frequent unique answers produced by the target model for each key:

\newcommand\userinput[1]{\textbf{#1}}

\begin{quote}\begin{scriptsize}\begin{verbatim}
Tax Number: {
    `91-0837469': 280,
    `56-0589582': 45,
    `91-0857469': 23,
    `91-08374': 12,
    `1828': 7,
    `91-0657469': 5,
    `37-1159433': 5,
    `orig cf': 3,
    `865864-ny': 3,
    `91-08574': 2,
    `033797': 1,
    `db829057-8ea2-4d34-abbd-ec53': 1,
    `woc13422932 [00.00]': 1,
    `92-346-369': 1,
    `113 61779 rt': 1
    ...
}
Registration ID: {
    `91-0837469': 30,
    `865864-ny': 11,
    `216-256-5304': 9,
    `91-08374': 8,
    `56-0589582': 8,
    `pol-cand': 8,
    `921-0850.00', 7,
    `56-05895', 6,
    `idb#1828', 6,
    `91-0857469', 5,
    `pol-iss', 5,
    `pb-18', 5,
    `ktvf', 5,
    `92-346-369': 1,
    `orig cf', 5
    ...
}
\end{verbatim}\end{scriptsize}\end{quote}

\newpage
Based on the results, we see highly skewed distributions over the predicted answer, while we expect more uniform ones. 
Interestingly, there are common answers between these two sets, suggesting that the model is confused between these two queries after being under-trained. We then run the test against \vtfz model and confirm that \textit{none} of the top predicted answers from \vtfc comes from the pre-training stage. 

By accessing to the training data, we further inspect among top-15 most predicted answers in each distribution, and find that: \textit{91-0837469} is the tax ID of \textit{airborne express}; \textit{56-0589582} is the tax ID of \textit{kennedy covington lobdell \& hickman. l.l.p}; \textit{idb\#1828} is the tax ID of \textit{wajq-fm}; \textit{37-1159433} is the tax ID of \textit{meyer, capel, hirschfeld, muncy, jahn \& aldeen, p.c.}, all the providers listed are included in training data. Also, we find that among the predicted answers in both sets, there are 103 answers which correspond to answers for different training questions, and they are not necessarily relevant to \textit{Tax Number/Registration ID}, but other information such as \textit{Phone Number/Asset Code} etc.
This clearly demonstrates that some sensitive information is actually memorized inside non-private models 

\subsection{Private Mechanisms mitigate Memorization}\label{sec:supp_memorization_test_dp}
We further validate the effectiveness of our DP mechanism to alleviate the memorization effect. First, we see that introducing DP to the training algorithm significantly closes the model's performance gap $\Delta$ between $\mathcal{D}_{in}$/$\mathcal{D}_{out}$ subsets, as shown in \cref{tab:supp_overfitting_dp}. Furthermore, we observe significant drops from Private Models compared to Non-Private ones in our \memtest with both \textit{Provider Name} and \textit{Email}, to almost 0, as reported in \cref{tab:supp_provider_name} and \cref{tab:supp_provider_email}.
These results suggest that overfitting in our Private Models is effectively decreased, and thus reduces memorization.

\vspace{-0.5cm}
\begin{table}[H]
\scriptsize
\centering
\begin{tabular}{|l|cc|cc|cc|c|c|}
\hline
\multicolumn{1}{|c|}{\multirow{2}{*}{Model}} & \multicolumn{2}{c|}{RED}           & \multicolumn{2}{c|}{RED Positive}  & \multicolumn{2}{c|}{RED Negative}  & \multirow{2}{*}{$\Delta$ACC} & \multirow{2}{*}{$\Delta$ANLS} \\ \cline{2-7}
                       & \multicolumn{1}{c|}{ACC}   & ANLS  & \multicolumn{1}{c|}{ACC}   & ANLS  & \multicolumn{1}{c|}{ACC}   & ANLS  &                              &                               \\ \hline
\vtfc                 & \multicolumn{1}{c|}{81.40} & 90.17 & \multicolumn{1}{c|}{85.92} & 93.68 & \multicolumn{1}{c|}{76.53} & 86.48 & \textbf{\textcolor{red}{9.39}} & \textbf{\textcolor{red}{7.20}}\\
\hline
\vtfcdp ({\small $\varepsilon$}=1)               & \multicolumn{1}{c|}{52.4} & 60.09 & \multicolumn{1}{c|}{53.98} & 61.87 & \multicolumn{1}{c|}{50.68} & 58.17 & 3.3 & 3.7\\
\vtfcdp ({\small $\varepsilon$}=4)                & \multicolumn{1}{c|}{58.59} & 65.77 & \multicolumn{1}{c|}{59.86} & 67.05 & \multicolumn{1}{c|}{57.23} & 64.39 & 2.63 & 2.66 \\
\vtfcdp ({\small $\varepsilon$}=8)               & \multicolumn{1}{c|}{60.31} & 67.44 & \multicolumn{1}{c|}{61.57} & 69.04 & \multicolumn{1}{c|}{58.94} & 65.72 & 2.62    & 3.33  \\
\hline
\vtffldp ({\small $\varepsilon$}=1)              & \multicolumn{1}{c|}{49.94} & 56.8 & \multicolumn{1}{c|}{50.96} & 58.23 & \multicolumn{1}{c|}{48.83} & 55.25 & 2.13 & 2.98\\ 
\vtffldp ({\small $\varepsilon$}=4)              & \multicolumn{1}{c|}{53.68} & 61.8 & \multicolumn{1}{c|}{55.01} & 63.21 & \multicolumn{1}{c|}{52.25} & 60.28 & 2.76 & 2.93\\ 
\vtffldp ({\small $\varepsilon$}=8)              & \multicolumn{1}{c|}{58.33}  &  65.44  & \multicolumn{1}{c|}{59.41} &  66.89 & \multicolumn{1}{c|}{57.18}  &  63.88 & 2.22  & 3.01 \\ 
\hline
\end{tabular}
\vspace{0.25cm}
\caption{\textbf{DocVQA Performance of Private Models on RED.} Compared to the Non-Private Model \vtfc, $\Delta$ values from private Models are significantly decreased. Refer to \cref{table:overfitting} for details of notations.}
\label{tab:supp_overfitting_dp}
\end{table}

\newpage
\subsection{More on Attack Performance against Private Models}\label{sec:supp_pmia_results_dp}
We detail in \cref{tab:pmia_results} the results of our proposed attacks against Private Models, both with centralized training and Federated Learning. %

\begin{table*}[ht]
\caption{\textbf{Attack Performance against Private and Non-Private VT5 models on different RED subsets.} Colored rows indicate experiments with Ensemble. ZK/PK denotes our \textit{zero-knowledge}/\textit{partial-knowledge} setting while U/S denotes the Unsupervised/Supervised model training, respectively. $T_{s}$ indicates the number of test providers in each evaluation subset, where all providers with fewer than $s+1$ test questions are eliminated. Train/Test indicates the size of $\mathcal{P}_{train}/\mathcal{P}_{test}$ in the corresponding approach, with $r=0.15$ is the sampling ratio used in APK. Results are reported with the standard deviation over 5 random seeds. }
\label{tab:pmia_results}
\centering
\scalebox{0.73}{
\begin{tabular}{|c|c|l|l|c|c|c|ccc|}
\hline
    & & & & & & & \multicolumn{3}{c|}{Evaluation Set $T_s$} \\
    
\cline{8-10}

\multirow{-2}{*}{Model}   & \multirow{-2}{*}{\Ebudget} & \multicolumn{1}{c|}{\multirow{-2}{*}{Attack}}   & \multicolumn{1}{c|}{\multirow{-2}{*}{Feature Set}} &
    \multirow{-2}{*}{Setting}      &    \multirow{-2}{*}{Train}        & \multirow{-2}{*}{Test}    &
    \multicolumn{1}{c|}{$s=0$}     &    \multicolumn{1}{c|}{$s=5$}     & $s=10$  \\ \hline

    & &                           & $G_1$         & ZK + U    & 0         & $(1-r)T_s$    & \multicolumn{1}{c|}{56.13$\pm$0.55} & \multicolumn{1}{c|}{60.25$\pm$0.62} & 65.12$\pm$0.99  \\
    & & \multirow{-2}{*}{AZK}     & $G_1$+$G_4$   & ZK + U    & 0         & $(1-r)T_s$    & \multicolumn{1}{c|}{48.25$\pm$5.67} & \multicolumn{1}{c|}{47.80$\pm$3.30} & 58.16$\pm$6.60  \\
    
\cline{3-10} 

    & &                       & $G_1$             & ZK + S   & $rT_s$   & $(1-r)T_s$    & \multicolumn{1}{c|}{55.11$\pm$2.20}  & \multicolumn{1}{c|}{58.23$\pm$2.67} & 60.29$\pm$3.15  \\
    & &                       & $G_1$+$G_2$+$G_3$ & PK + S   & $rT_s$   & $(1-r)T_s$    & \multicolumn{1}{c|}{59.67$\pm$1.01}  & \multicolumn{1}{c|}{{\color[HTML]{3531FF} \textbf{61.66$\pm$1.71}}} & {\color[HTML]{3531FF} \textbf{65.20$\pm$1.81}} \\

    & & \multirow{-3}{*}{APK} & $G_1$+$G_2$+$G_3$+$G_4$ & PK + S & $rT_s$ & $(1-r)T_s$ & \multicolumn{1}{c|}{{\color[HTML]{3531FF} \textbf{60.97$\pm$1.40}}}   & \multicolumn{1}{c|}{61.26$\pm$1.57} & 64.44$\pm$2.62 \\
    
\cline{3-10} 

    & & \cellcolor[HTML]{DAE8FC}AZK+$\text{CLS}_{1}$+$\text{CLS}_{2}$ & \cellcolor[HTML]{DAE8FC}$G_1$+$G_4$ & \cellcolor[HTML]{DAE8FC}ZK + U & \cellcolor[HTML]{DAE8FC}0      & \cellcolor[HTML]{DAE8FC}$(1-r)T_s$ & \multicolumn{1}{c|}{\cellcolor[HTML]{DAE8FC}60.56$\pm$0.70}           & \multicolumn{1}{c|}{\cellcolor[HTML]{DAE8FC}\textbf{63.68$\pm$0.80}} & \cellcolor[HTML]{DAE8FC}66.25$\pm$1.80         \\

\multirow{-7}{*}{\vtfc} & \multirow{-7}{*}{$\infty$} & \cellcolor[HTML]{DAE8FC}APK+$\text{CLS}_{1}$+$\text{CLS}_{2}$ & \cellcolor[HTML]{DAE8FC}$G_1$+$G_2$+$G_3$+$G_4$    & \cellcolor[HTML]{DAE8FC}PK + S & \cellcolor[HTML]{DAE8FC}$rT_s$ & \cellcolor[HTML]{DAE8FC}$(1-r)T_s$ & \multicolumn{1}{c|}{\cellcolor[HTML]{DAE8FC}\textbf{61.84$\pm$0.09}} & \multicolumn{1}{c|}{\cellcolor[HTML]{DAE8FC}62.53$\pm$0.00}            & \cellcolor[HTML]{DAE8FC}\textbf{66.99$\pm$0.00}  \\ 

\hline
    & & AZK   & $G_1$ & ZK + U    & 0 & $(1-r)T_s$    & \multicolumn{1}{c|}{55.12$\pm$0.85}   & \multicolumn{1}{c|}{58.54$\pm$0.77}  & 60.77$\pm$0.94 \\
    
  & \multirow{-2}{*}{8} & APK & $G_1$+$G_2$+$G_3$    & PK + S    & $rT_s$    & $(1-r)T_s$    & \multicolumn{1}{c|}{51.67$\pm$1.56} & \multicolumn{1}{c|}{54.06$\pm$2.54}     & 58.55$\pm$1.34 \\

\cline{2-10} 
    & & AZK & $G_1$ & ZK + U & 0 & $(1-r)T_s$ & \multicolumn{1}{c|}{53.59$\pm$0.76} & \multicolumn{1}{c|}{56.67$\pm$0.69} & 59.03$\pm$1.53 \\
    
\multirow{-2}{*}{\vtfcdp}  & \multirow{-2}{*}{4} & APK & $G_1$+$G_2$+$G_3$    & PK + S    & $rT_s$    & $(1-r)T_s$    & \multicolumn{1}{c|}{54.20$\pm$0.98} & \multicolumn{1}{c|}{53.45$\pm$3.05}     & 54.88$\pm$2.94 \\

\cline{2-10} 

    & & AZK & $G_1$ & ZK + U & 0 & $(1-r)T_s$ & \multicolumn{1}{c|}{55.37$\pm$3.06} & \multicolumn{1}{c|}{57.17$\pm$0.73} & 60.48$\pm$1.48 \\
    
   & \multirow{-2}{*}{1} & APK & $G_1$+$G_2$+$G_3$    & PK + S    & $rT_s$    & $(1-r)T_s$    & \multicolumn{1}{c|}{51.71$\pm$3.17} & \multicolumn{1}{c|}{50.27$\pm$2.14}     & 53.04$\pm$0.64 \\

\hline
    & & AZK & $G_1$ & ZK + U & 0 & $(1-r)T_s$ & \multicolumn{1}{c|}{53.95$\pm$0.72} & \multicolumn{1}{c|}{57.88$\pm$0.70}     & 61.35$\pm$0.43 \\
    
   \multirow{-2}{*}{\vtffl} & \multirow{-2}{*}{$\infty$} & APK & $G_1$+$G_2$+$G_3$    & PK + S    & $rT_s$    & $(1-r)T_s$    & \multicolumn{1}{c|}{55.57$\pm$2.48} & \multicolumn{1}{c|}{58.72$\pm$2.02}     & 62.13$\pm$2.71 \\
   
\hline
    & & AZK   & $G_1$ & ZK + U    & 0 & $(1-r)T_s$    & \multicolumn{1}{c|}{53.81$\pm$0.97} & \multicolumn{1}{c|}{57.78$\pm$0.94}     & 61.54$\pm$0.78 \\
    
  & \multirow{-2}{*}{8} & APK & $G_1$+$G_2$+$G_3$    & PK + S    & $rT_s$    & $(1-r)T_s$   & \multicolumn{1}{c|}{52.56$\pm$2.05} & \multicolumn{1}{c|}{52.70$\pm$4.21}     & 56.43$\pm$1.45 \\

\cline{2-10} 
    & & AZK & $G_1$ & ZK + U & 0 & $(1-r)T_s$ & \multicolumn{1}{c|}{53.34$\pm$0.96} & \multicolumn{1}{c|}{55.82$\pm$0.79}     & 61.35$\pm$0.86 \\
    
\multirow{-2}{*}{\vtffldp}  & \multirow{-2}{*}{4} & APK & $G_1$+$G_2$+$G_3$    & PK + S    & $rT_s$    & $(1-r)T_s$    & \multicolumn{1}{c|}{52.45$\pm$1.35} & \multicolumn{1}{c|}{51.28$\pm$1.99}     & 56.91$\pm$4.40 \\

\cline{2-10} 

    & & AZK & $G_1$ & ZK + U & 0 & $(1-r)T_s$ & \multicolumn{1}{c|}{56.30$\pm$0.72} & \multicolumn{1}{c|}{57.48$\pm$0.65}     & 60.00$\pm$2.65 \\
    
   & \multirow{-2}{*}{1} & APK & $G_1$+$G_2$+$G_3$    & PK + S    & $rT_s$    & $(1-r)T_s$    & \multicolumn{1}{c|}{52.17$\pm$2.12} & \multicolumn{1}{c|}{53.55$\pm$4.94}     & 54.69$\pm$3.93 \\
\hline

\end{tabular}
}
\end{table*}

\newpage
\subsection{Ablation Study}\label{sec:supp_pmia_ablation}
In this section, we measure the impact of our selected metric and classifying methods for both AZK and APK %
in terms of Attack Accuracy. We employ \vtfc as the target model and perform hyperparameter search for Random Forest in all ablation experiments. 
\subsubsection{Selected Metrics and Classifying Methods}
We ablate each of the selected metrics according to its availability order in our attack scenario, i.e. from \textit{partial-knowledge} with access to the pre-trained model and loss/confidence values to \textit{zero-knowledge}. We evaluate all the attacks on the subset with $s=5$ and report the results in \cref{tab:supp_ablate_metrics}.
From line 1 to 4, we find that K-Means method is a good baseline in restricted scenario like \textit{zero-knowledge}, where only ACC and NLS are available metrics. In contrast, Random Forest is not suitable choice for this setting as it requires some training data and more features to perform adequately.
When more information is accessible like in \textit{partial-knowledge} scenario, the Supervised training directly benefits from it while the Unsupervised one shows substantial degradation with higher dimensional features (line 7). 
Finally, the designed features are all useful for Supervised training (line 2 to 6), as the model incrementally improves and surpasses the best performance of Unsupervised approach with the full set of features.
\begin{table}[h]
\centering
\begin{tabular}{ccccccccc}
\hline\noalign{\smallskip}
& \multirow{2}{*}{Setting} & \multicolumn{6}{c}{Feature} & \multirow{2}{*}{Accuracy}\\
\cline{3-8}
\noalign{\smallskip}
&  & ACC & NLS & $\mathit{L}$ & $\mathit{\conf}$ & $\Delta{L}$ & $\Delta{\conf}$ & \\
\noalign{\smallskip}
\hline
\noalign{\smallskip}	
\multirow{2}{*}{AZK} & KM+U & $\checkmark$ & & & & & & 60.15$\pm$0.43\\
 & \cellcolor[HTML]{EFEFEF}{KM+U} & \cellcolor[HTML]{EFEFEF}{$\checkmark$} & \cellcolor[HTML]{EFEFEF}{$\checkmark$} & \cellcolor[HTML]{EFEFEF}{} & \cellcolor[HTML]{EFEFEF}{} & \cellcolor[HTML]{EFEFEF}{} & \cellcolor[HTML]{EFEFEF}{} & \cellcolor[HTML]{EFEFEF}{60.25$\pm$0.62} \\
\noalign{\smallskip}
\hline
\noalign{\smallskip}						
\multirow{5}{*}{APK} & RF+S & $\checkmark$ & & & & & & 57.23$\pm$1.36 \\
 & RF+S & $\checkmark$ & $\checkmark$ & & & & & 58.24$\pm$2.68 \\
 & RF+S & $\checkmark$ & $\checkmark$ & $\checkmark$ & $\checkmark$ & & & 59.79$\pm$1.75\\
 & \cellcolor[HTML]{EFEFEF}{RF+S} & \cellcolor[HTML]{EFEFEF}{$\checkmark$} & \cellcolor[HTML]{EFEFEF}{$\checkmark$} & \cellcolor[HTML]{EFEFEF}{$\checkmark$} & \cellcolor[HTML]{EFEFEF}{$\checkmark$} & \cellcolor[HTML]{EFEFEF}{$\checkmark$} & \cellcolor[HTML]{EFEFEF}{$\checkmark$} & \cellcolor[HTML]{EFEFEF}{61.66$\pm$1.71}\\
& KM+U & $\checkmark$ & $\checkmark$ & $\checkmark$ & $\checkmark$ & $\checkmark$ & $\checkmark$ & 49.22$\pm$2.3 \\ 
\noalign{\smallskip}
\hline
\end{tabular}
\vspace{0.25cm}
\caption{\textbf{Ablation Study of the Selected Metrics and Classifying Methods for our proposed attacks}. KM denotes K-Means Clustering method while RF denotes Random Forest Classifier. In all experiments, the performance is evaluated on the same RED subset with size $(1-r)T_s$, where we use $s=5$ and $r=0.15$ for Supervised Setting. Refer to \cref{tab:pmia_results} for details of notation.}
\label{tab:supp_ablate_metrics}
\end{table}

\newpage
\subsubsection{Minimum Number of Queries $s$}
\begin{figure}[H]
\centering

\includegraphics[width=0.90\textwidth]{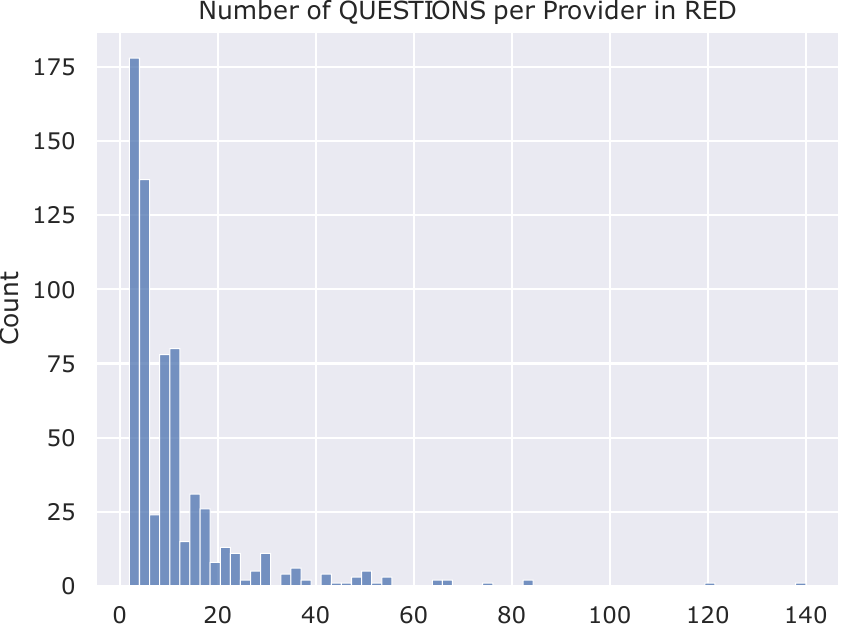}
\label{fig:num_question_red}

  \vspace{0.8cm}
  
  \centering
    \begin{tabular}{cccccc}
        \hline\noalign{\smallskip}
        $s$ & 0 (All) & 5 & 10 & 15 & 20 \\
        \noalign{\smallskip}
        \hline
        \noalign{\smallskip}
        $N\textsubscript{Provider}$ & 660 & 466 & 243 & 122 & 83\\
        APK$\ddag$ & 59.67$\pm$1.01 & 61.66$\pm$1.71 & \textbf{65.2$\pm$1.81} & 61.57$\pm$4.18 & 64.05$\pm$1.83\\
        \noalign{\smallskip}
        \hline
        \end{tabular}
\caption{\textbf{(Top) Distribution of number of Questions per Provider} in the RED data. \textbf{(Bottom) The impact of the number of Questions per Provider} on performance of our proposed attacks. $N\textsubscript{Provider}$ is the total number of Train/Test Providers with the corresponding $s$.}
\label{fig:supp_ablate_num_query}
\end{figure}

We also evaluate the performance of $APK\ddag$ while varying the minimum number of Questions per Provider $s$. As illustrated in Tab.~\ref{fig:supp_ablate_num_query}, there is an upward trend in Attack Accuracy when $s$ is increased from 0 to 10. 
This is due to the retention of informative providers, filtering out outliers and thus creating a more representative training dataset. However, excessively raising $s$ as the threshold has an adverse effect, as it reduces the total number of Providers.

\newpage
\section{Training hyperparameters}
In \cref{tab:fl_methods_hyperparams} we specify the hyperparameters used to train the different methods.

\begin{table}[ht]
\centering
\scriptsize
\adjustbox{max width=\textwidth}{
\begin{tabularx}{1.08\linewidth}{p{0.18\linewidth} P{0.1\linewidth} P{0.1\linewidth} P{0.11\linewidth} P{0.1\linewidth} P{0.1\linewidth} P{0.18\linewidth} P{0.15\linewidth}} \toprule
\multirow{2}{*}{Method} & Learning  & Batch & Epochs /  & \multirow{2}{*}{$\delta$} & Sensitivity   & Noise         & Providers      \\
                        & Rate      & Size  & Iterations    &                           & $\sigma$      & Multiplier    & per iteration        \\
\midrule
\vtfc                  & $2e^{-4}$ & 10 & 10 & -           & - & -                 & 4149 (All) \\
\vtfcdp~\Ebudget=$8$   & $2e^{-4}$ & 8  & 10 & $10e^{-5}$  & 5.0 & $0.83251953125$   & 1000   \\
\vtfcdp~\Ebudget=$4$   & $2e^{-4}$ & 8  & 10 & $10e^{-5}$  & 2.5 & $1.25244140625$   & 1000   \\
\vtfcdp~\Ebudget=$1$   & $2e^{-4}$ & 8  & 10 & $10e^{-5}$  & 1.0 & $3.32031250000$   & 1000   \\
\bottomrule
\end{tabularx}}
\vspace{0.40cm}
\\
\adjustbox{max width=\textwidth}{
\begin{tabularx}{1.25\linewidth}{p{0.15\linewidth} P{0.1\linewidth} P{0.1\linewidth} P{0.1\linewidth} P{0.1\linewidth} P{0.1\linewidth} P{0.15\linewidth} P{0.19\linewidth} P{0.19\linewidth}} \toprule
\multirow{2}{*}{Method} & Learning  & Batch & FL        & \multirow{2}{*}{$\delta$} & Sensitivity   & Noise         & Provider sampling & Client sampling    \\
                        & Rate      & Size  & Rounds    &                           & $\sigma$      & Multiplier    & probability       & probability \\
\midrule
\vtffl                  & $2e^{-4}$ & 10 & 10 & -           & - & -                 & 1 & 0.2 \\
\vtffldp~\Ebudget=$8$   & $2e^{-4}$ & 8  & 10 & $10e^{-5}$  & 5.0 & $0.77148437500$ & 1 & 0.2 \\
\vtffldp~\Ebudget=$4$   & $2e^{-4}$ & 8  & 10 & $10e^{-5}$  & 5.0 & $1.13647460937$ & 1 & 0.2 \\
\vtffldp~\Ebudget=$1$   & $2e^{-4}$ & 8  & 10 & $10e^{-5}$  & 1.0 & $2.85156250000$ & 1 & 0.2 \\
\bottomrule
\end{tabularx}
}
\caption{\textbf{Training hyperparameters} for the different centralized (top) and federated learning (bottom) methods used in the \datasetName dataset.}
\label{tab:fl_methods_hyperparams}
\end{table}

\section{Privacy Analysis: More details}
\label{sec:priv_analysis_appendix}

FL-PROVIDER-DP is described in Alg.~\ref{alg:group_fed_learn}.
Group-level differential privacy necessitates that each client adds Gaussian noise to the aggregated model updates, which are generated using the data from the providers. We define $\mathbb{P}_k$ as a set of predefined disjoint providers in the dataset of client $k$. In particular, each client first select $\mathbb{M} \subseteq \mathbb{P}_k$ randomly. Then, for each selected provider $i$ in $\mathbb{M}$, we 
compute the update $\Delta \mbf{w}_t^{i,k}= \mathbf{AdamW}(i, \mbf{w}', \Tgd) - \mbf{w}_{t-1}$, 
which is then clipped (in Line 17) to obtain $\Delta \hat{\mbf{w}}_t^{i,k}$ with $L_2$-norm at most $C$. Then, random noise $\mbf{z}_{k} \sim \mathcal{N}(0,\sigma^2 C^2 \mathbf{I}/|\mathbb{K}|)$ is added to $\sum_{i \in \mathbb{M}}  \Delta \hat{\mbf{w}}_t^{i,k}$ before averaging over $M$ to obtain the update $\mathbf{u}_{t}^{k}= \frac{1}{M} \left( \sum_{i \in \mathbb{M}}  \Delta \hat{\mbf{w}}_t^{i,k} + \mbf{z}_{k} \right )$ for client $k$. Then, we compute at the server side the aggregate $\sum_{k \in \mathbb{K}} \mathbf{u}_{t}^{k}= \sum_{k \in \mathbb{K}} \left( \frac{1}{M} \left( \sum_{i \in \mathbb{M}}  \Delta \hat{\mbf{w}}_t^{i,k} + \mbf{z}_{k} \right ) \right )  =  \sum_{k \in \mathbb{K}} \left( \frac{1}{M} \left( \sum_{i \in \mathbb{M}}  \Delta \hat{\mbf{w}}_t^{i,k} \right ) \right ) + \mathcal{N}(0,\frac{\sigma^2 C^2 \mathbf{I}}{M^2})$ as the sum of Gaussian random variables also follows Gaussian distribution\footnote{More precisely, $\sum_{i}\mathcal{N}(\nu_{i},\xi_{i})=\mathcal{N}(\sum_{i} \nu_{i},\sqrt{\sum_{i}\xi_{i}^{2}})$}:\\

\begin{align}
    \sum_{k\in\mathbb{K}} \mathbf{u}_t^k &= \sum_{k} \frac{1}{M_k} \left( \sum_{i \in \mathbb{M}}  \Delta \hat{\mbf{w}}_t^{i,k}
    + \mathcal{N}(0, \frac{C^2\sigma^2 I}{|\mathbb{K}|})  \right) \nonumber \\
    &= \sum_{k} \left( \sum_{i \in \mathbb{M}} \frac{  \Delta \hat{\mbf{w}}_t^{i,k}}{M_k}
    + \frac{1}{M_k} \mathcal{N}(0, \frac{C^2\sigma^2 I}{|\mathbb{K}|}) \right) \nonumber \\
    &= \sum_{k} \sum_{i \in \mathbb{M}} \frac{\Delta \hat{\mbf{w}}_t^{i,k}}{M_k}
    + \sum_{k} \frac{1}{M_k} \mathcal{N}(0, \frac{C^2\sigma^2 I}{|\mathbb{K}|})  \nonumber \\
    &= \sum_{k} \sum_{i \in \mathbb{M}} \frac{\Delta \hat{\mbf{w}}_t^{i,k}}{M_k}
    + \mathcal{N}\left(0, \sum_{k} \frac{C^2\sigma^2}{|\mathbb{K}|M_k^2} I\right) \nonumber
\end{align}

Assuming $M_k = M$ for all $k$:

\begin{align}
    &= \sum_{k} \sum_{i \in \mathbb{M}} \frac{\Delta \hat{\mbf{w}}_t^{i,k}}{M}
    + \mathcal{N}\left(0, \sum_k \frac{C^2\sigma^2}{|\mathbb{K}|M^2} I\right) \nonumber \\
    &= \sum_{k} \sum_{i \in \mathbb{M}} \frac{\Delta \hat{\mbf{w}}_t^{i,k}}{M}
    + \mathcal{N}\left(0, \frac{|\mathbb{K}| C^2\sigma^2}{|\mathbb{K}|M^2} I\right) \nonumber
\end{align}

And then differential privacy is satisfied where $\varepsilon$ and $\delta$ can be computed using the PRV accountant described in Section \ref{sec:dpfl}.

However, as the noise is inversely proportional to $|\mathbb{K}|$, 
$\mbf{z}_{k}$ is likely to be small if $|\mathbb{K}|$ is too large. Therefore, the adversary accessing an individual update $\mathbf{u}_{t}^{k}$ can almost learn a non-noisy update since $\mbf{z}_{k}$ is small. Hence, each client uses secure aggregation to encrypt its individual update before sending it to the server. Upon reception, the server sums the encrypted updates as:
\begin{align}
\sum_{k\in\mathbb{K}} \mathbf{u}_t^k &= \sum_{k\in\mathbb{K}}\mathsf{Enc}_{K_k} \left( \frac{1}{M} \left( \sum_{i \in \mathbb{M}}  \Delta \hat{\mbf{w}}_t^{i,k} + \mbf{z}_{k} \right ) \right ) \nonumber \\
& = \sum_{k \in \mathbb{K}} \left( \frac{1}{M} \left( \sum_{i \in \mathbb{M}}  \Delta \hat{\mbf{w}}_t^{i,k} \right ) \right ) + \mathcal{N}(0,\frac{\sigma^2 C^2 \mathbf{I}}{M^2}) \label{eq:a1}
\end{align}
where $\mathsf{Enc}_{K_k}\left(\frac{1}{M} \left( \sum_{i \in \mathbb{M}}  \Delta \hat{\mbf{w}}_t^{i,k} + \mbf{z}_{k} \right ) \right ) = \frac{1}{M} \left( \sum_{i \in \mathbb{M}}  \Delta \hat{\mbf{w}}_t^{i,k} + \mbf{z}_{k} \right ) + \mbf{K}_k \mod p$ and $\sum_{k}\mbf{K}_k=0$ (see \cite{AcsC11,bonawitz2016practical} for more details). Here the modulo is taken element-wise and $p=2^{\lceil \log_{2}(\max_{k}||\frac{1}{M} \left( \sum_{i \in \mathbb{M}}  \Delta \hat{\mbf{w}}_t^{i,k} + \mbf{z}_{k} \right )||_{\infty}|\mathbb{K}|)\rceil}$. \\

\end{document}